\documentclass{article}

\usepackage{microtype}
\usepackage{graphicx}
\usepackage{booktabs} 

\usepackage{hyperref}

\usepackage[accepted]{icml2024}

\usepackage{amsmath}
\usepackage{amssymb}
\usepackage{mathtools}
\usepackage{amsthm}

\usepackage[capitalize,noabbrev]{cleveref}

\theoremstyle{plain}

\theoremstyle{definition}

\theoremstyle{remark}

\usepackage[textsize=tiny]{todonotes}

\usepackage{amsmath,amsfonts,bm}

\def\eqref#1{equation~\ref{#1}}

\def\1{\bm{1}}

\DeclareMathAlphabet{\mathsfit}{\encodingdefault}{\sfdefault}{m}{sl}
\SetMathAlphabet{\mathsfit}{bold}{\encodingdefault}{\sfdefault}{bx}{n}

\usepackage{url}
\usepackage{footmisc}
\usepackage{caption}
\usepackage{subcaption}
\usepackage{xcolor}
\usepackage{tikz}
\usetikzlibrary{bayesnet}
\usepackage{color}
\usetikzlibrary{positioning, arrows, automata}
\usetikzlibrary{backgrounds, calc}
\usepackage{pgfplots}
\pgfplotsset{compat=1.8}
\usepackage{pgfplotstable}
\usepgfplotslibrary{groupplots}

\newcommand{\cvec}[1]{\boldsymbol{\mathrm{#1}}}

\newcommand{\st}{\textrm{s.t.}}
\newcommand{\old}[1]{{#1}_{\textrm{old}}}

\newcommand{\context}[0]{\cvec c}
\newcommand{\option}[0]{o}
\newcommand{\mpparam}[0]{\cvec {\theta}}
\newcommand{\KL}[2]{\textrm{KL}\left( {#1} \parallel {#2} \right)}

\newcommand{\joint}[1]{{#1}(\context, \mpparam)}
\newcommand{\policy}[1]{{#1}(\mpparam |\context)}

\newcommand{\responsibility}[1]{{#1}(\option|\context, \mpparam)}
\newcommand{\tilderesponsibility}[1]{\Tilde{#1}(o|\context, \mpparam)}

\newcommand{\gating}[1]{{#1}(\option|\context)}
\newcommand{\mgating}[1]{{#1}(\option)}
\newcommand{\tildegating}[1]{\Tilde{#1}(\option|\context)}

\newcommand{\contextdistr}[1]{{#1}\left(\context\right)}

\newcommand{\contextcomp}[1]{{#1}(\context|\option)}

\newcommand{\potential}[2]{{\cvec{#1}_{#2}}(\context)}
\newcommand{\normalizer}[0]{Z}

\newcommand{\expert}[1]{{#1}(\mpparam | \context, \option)}

\newcommand{\reward}[0]{\textrm{R}(\context, \mpparam)}

\icmltitlerunning{Acquiring Diverse Skills using Curriculum Reinforcement Learning with Mixture of Experts}

\begin{document}

\twocolumn[
\icmltitle{Acquiring Diverse Skills using Curriculum Reinforcement Learning with Mixture of Experts}

\icmlsetsymbol{equal}{*}

\begin{icmlauthorlist}
\icmlauthor{Onur Celik}{kit,fzi}
\icmlauthor{Aleksandar Taranovic}{kit}
\icmlauthor{Gerhard Neumann}{kit,fzi}
\end{icmlauthorlist}

\icmlaffiliation{kit}{Autonomous Learning Robots, Karlsruhe Institute of Technology, Karlsruhe, Germany}
\icmlaffiliation{fzi}{FZI Research Center for Information Technology, Karlsruhe, Germany}

\icmlcorrespondingauthor{Onur Celik}{celik@kit.edu}

\icmlkeywords{Reinforcement Learning, Diverse Skill Learning, Skill Discovery, Automatic Curriculum Learning, Mixture of Experts, Machine Learning, ICML}

\vskip 0.3in
]

\printAffiliationsAndNotice{}  
\begin{abstract}
Reinforcement learning (RL) is a powerful approach for acquiring a good-performing policy. However, learning diverse skills is challenging in RL due to the commonly used Gaussian policy parameterization. We propose \textbf{Di}verse \textbf{Skil}l \textbf{L}earning (Di-SkilL\footnote{Videos and code are available on the project webpage: \url{https://alrhub.github.io/di-skill-website/}}), an RL method for learning diverse skills using Mixture of Experts, where each expert formalizes a skill as a contextual motion primitive. Di-SkilL optimizes each expert and its associate context distribution to a maximum entropy objective that incentivizes learning diverse skills in similar contexts. The per-expert context distribution enables automatic curricula learning, allowing each expert to focus on its best-performing sub-region of the context space. To overcome hard discontinuities and multi-modalities without any prior knowledge of the environment's unknown context probability space, we leverage energy-based models to represent the per-expert context distributions and demonstrate how we can efficiently train them using the standard policy gradient objective. We show on challenging robot simulation tasks that Di-SkilL can learn diverse and performant skills. 
\end{abstract}

\section{Introduction}

Solving tasks in diverse manners enables agents to better adapt to unknown and challenging situations. This diverse skill set is beneficial in many scenarios, such as playing table tennis, where applying different strikes (e.g. backhand, forehand, or smashing) to similar incoming balls is advantageous because the strike is less predictable for the opponent. Similarly, in scenarios with environmental changes where learned skills might be infeasible over time (e.g. grasping an object while avoiding obstacles), diverse skills provide additional adaptivity by discarding these invalid skills and relying on alternatives. This property makes them superior because complete relearning of skills is avoided.

Acquiring these diverse skill sets requires learning a policy that can represent multi-modality in the behavior space. Recent advances in supervised policy learning have demonstrated the potential of training high-capacity policies capable of capturing multi-modal behaviors \citep{shafiullah2022behavior, blessing2023information, chi2023diffusionpolicy, jia2023towards}. These policies exhibit remarkably diverse skills and outperform state-of-the-art methods. However, Reinforcement Learning (RL) is essential to acquire skills in cases where no expert data is available, or data collection is expensive. Discovering multi-modal behaviors using RL is challenging since the policies usually rely on Gaussian parameterization and thus can only discover a single behavior.  

We consider training agents that possess diverse skills, from which they can select to tackle a specific task differently. 
For capturing these multi-modalities in the agent's behavior space, we employ highly non-linear Mixture of Experts policies. Furthermore, we use automatic curriculum learning for efficient learning, enabling each expert to focus on a specific sub-region of the context space it favors. We introduce this curriculum shaping by optimizing for an additional per-expert context distribution that is used to sample contexts from the preferred regions to train the corresponding expert. Automatic curriculum learning has proven to increase performance by improving the exploration of agents, particularly in sparse-rewarded environments \citep{klink2022curriculum}. 
\begin{figure*}[t!]
    \begin{minipage}[b]{1\textwidth}
    \centering
       \resizebox{0.9\textwidth}{!}{\input{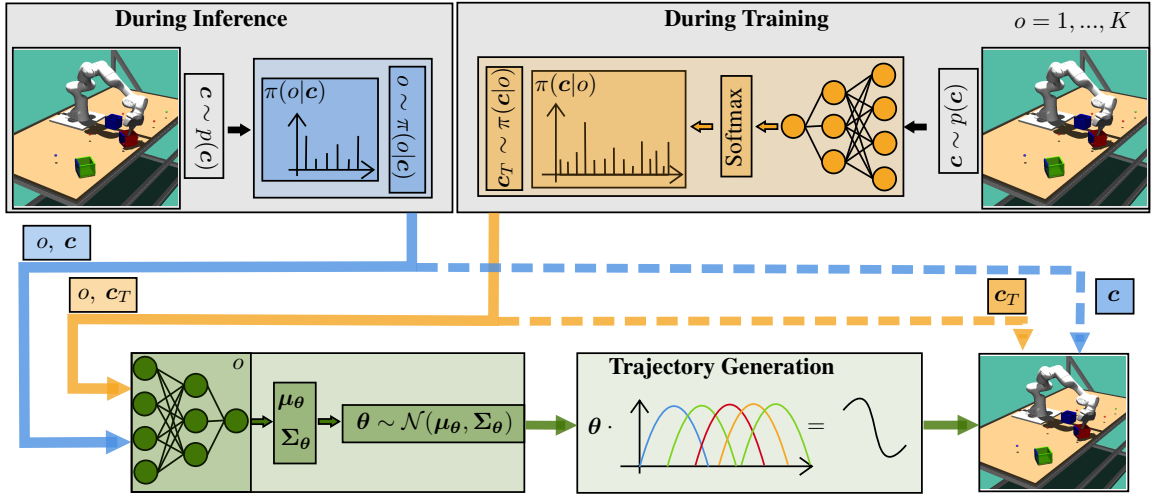}}
   \end{minipage}\hfill
   \caption{\textbf{The Sampling Procedure for Di-SkilL.} During \textbf{Inference} the agent observes contexts $\context$ from the environment's unknown context distribution $\contextdistr{p}$. The agent calculates the gating probabilities $\gating{\pi}$ for each context and samples an expert $\option$ resulting in $(\option, \context$) samples marked in blue. During \textbf{Training} we first sample a batch of contexts $\context$ from $\contextdistr{p}$, which is used to calculate the per-expert context distribution $\contextcomp{\pi}$ for each expert $\option = 1,..., K$. The $\contextcomp{\pi}$ provides a higher probability for contexts preferred by the expert $\expert{\pi}$. To enable curriculum learning, we provide each expert the contexts sampled from its corresponding $\contextcomp{\pi}$, resulting in the samples $(\option, \context_T)$ marked in orange. In both cases, the chosen $\expert{\pi}$ samples motion primitive parameters $\mpparam$ for each context, resulting in a trajectory $\tau$ that is subsequently executed on the environment. Before execution, the corresponding context, e.g., the goal position of a box, needs to be set in the environment. This is illustrated by the dashed arrows, with the context in blue for inference and orange for training.}
   \label{fig::overview}
\end{figure*}

We explore Contextual Reinforcement Learning in which a continuous-valued context describes the task \citep{kupcsik2013data}. In the example of robot table tennis (see Fig. \ref{fig::exps::envs}), a context includes the desired ball landing positions on the opponent's tableside as well as physical aspects, such as the incoming ball's velocity or friction properties. In continuous context spaces, the curriculum shaping per-expert context distributions are often parameterized as Gaussian \citep{klink2020self, celik2022specializing}. However, the agent is usually unaware of the context bounds, which makes additional techniques necessary to constrain the distribution updates to stay within the context region \citep{celik2022specializing}. Instead, we employ energy-based per-expert context distributions, which can be evaluated for any context and effectively represent multi-modality in the context space. Importantly, our model is trained solely using context samples from the environment that are inherently valid. Our approach eliminates the need for additional regularization of the context distribution and does not require prior knowledge about the environment. Due to the overlapping probability distributions of different per-expert contexts, our resulting mixture policy offers diverse solutions for similar contexts with a high probability. 

Recent research in RL has explored Mixture of Experts policies, but often these methods either train the mixture in unsupervised RL settings and then select the best-performing expert in the downstream task \citep{urlb, eysenbach2018diversity} or train linear experts, limiting their performance \citep{daniel2012hierarchical, celik2022specializing}. Our inspiration draws from recent advancements that have achieved diverse skill learning with a similar objective. However, their approach involves linear expert models with Gaussian context distributions. It requires prior knowledge of the environment to design a penalty term when the algorithm samples contexts outside the environment's bounds. These factors restrict the algorithm's performance and applicability when defining the context bounds requires knowledge, such as forward kinematics in robotics.

To summarize, we introduce a novel RL method for learning a Mixture of Experts policy that we refer to as \textbf{Di-SkilL} -- \textbf{Di}verse \textbf{Skil}l \textbf{L}earning (see Fig. \ref{fig::overview}). Our method can generalize to the continuous range of contexts defined by the (unknown) environment's context distribution while learning diverse, and non-linear skills for solving a task defined by a specific context. Importantly, our approach operates without any assumptions about the environment. We show how we can learn multi-modal context distributions by training an energy-based model solely on context samples obtained from the environment. On multiple sophisticated simulated robot tasks, we demonstrate that we can learn diverse skills while performing on par or better than baselines.

\section{Preliminaries}\label{sec::Preliminaries}
\textbf{Contextual Episode-based Policy Search (CEPS).} We consider learning diverse skills in the CEPS framework in which the continuous-valued context $\context \in \mathcal{C}$ defines the task, e.g. a goal location to reach. The context $\context \sim \contextdistr{p}$ is observed from the agent and is drawn from the environment's unknown context distribution $\contextdistr{p}$ at the beginning of each episode. The agent's search distribution $\policy{\pi}$ maps the context $\context$ to continuous-valued controller parameters $\mpparam \in \Theta$, which we represent as motion primitives (MP) \citep{schaal2006dynamic, paraschos2013probabilistic, li2023prodmp} (see Appendix \ref{sec::Appendix::additional_information}). We denote $\policy{\pi}$ as the agent's policy as common in the literature and optimize it by maximizing the objective 
\begin{align}
    \max_{\policy{\pi}} \mathbb{E}_{\contextdistr{p}}\left[\mathbb{E}_{\policy{\pi}}[\reward ] \right],
\end{align}
where $\reward$ denotes the return of a whole episode after executing the MP parameter $\mpparam$ in context $\context$. Due to the direct return optimization, CEPS does not require the Markov assumption as in common MDPs and is therefore specifically suitable for tasks where the formulation of a Markovian reward function is difficult.

\textbf{Mixture of Experts (MoE) Policy for Curriculum Learning.} Due to their ability to represent multi-modality, MoE policies are a favorable choice in diverse skill learning. The common MoE policy $\policy{\pi} = \sum_o \gating{\pi}\expert{\pi}$ \cite{bishop2006pattern} contains the gating distribution $\gating{\pi}$ that is assigning probabilities to each expert $o$ given context $\context$ during inference. However, to enable automatic curriculum learning during training, a learnable distribution $\contextdistr{\pi} = \sum_o \contextcomp{\pi}\mgating{\pi}$ is required that can explicitly choose and set context samples in the environment, so each expert $\option$ can decide on which contexts it favors training \cite{celik2022specializing}. Using Bayes' rule $\gating{\pi} = \contextcomp{\pi}\mgating{\pi}/\contextdistr{\pi}$ the MoE is rewritten as 
\begin{align}\label{eq::MixtureModel}
    \policy{\pi} = \sum_o \frac{\contextcomp{\pi} \mgating{\pi}}{\contextdistr{\pi}}\expert{\pi}.
\end{align}
The per-expert context distribution $\contextcomp{\pi}$ can now be optimized and allows the expert $o$ to choose contexts $\context$ it favors. We model each $\contextcomp{\pi}$ as an energy-based model and each $\expert{\pi}$ as a neural network returning a Gaussian distribution for a context $\context$ (see Fig. \ref{fig::overview} and Appendix \ref{sec::Appendix::additional_information}). The prior $\mgating{\pi}$ is set to a uniform distribution throughout this work.

\textbf{Self-Paced Diverse Skill Learning with MoE.} Due to its ability to represent multi-modality and automatic curriculum learning, the MoE model in Eq. \ref{eq::MixtureModel} is a suitable policy representation for discovering diverse skills in the same context-defined task. For explicit optimization of this policy, we are using the KL-regularized Maximum Entropy RL objective in CEPS \cite{celik2022specializing}
\begin{align}
\label{eq::maxEntrObj}
    \max_{\policy{\pi},\contextdistr{\pi}} & \mathbb{E}_{\contextdistr{\pi}}\left[\mathbb{E}_{\policy{\pi}} \left[\reward\right] + \alpha \textrm{H}\left[\policy{\pi}\right]\right] \nonumber\\
    &-\beta\KL{\contextdistr{\pi}}{\contextdistr{p}}.
\end{align}
The KL-term incentivizes the context distribution $\contextdistr{\pi}$ to match the environment's distribution $\contextdistr{p}$ and can be prioritized during optimization by choosing the scaling parameter $\beta$ appropriately. The entropy of the mixture model incentivizes learning diverse solutions \citep{celik2022specializing} and can be prioritized with a high scaling parameter $\alpha$. 
It is well-known that this objective is difficult to optimize for MoE policies and requires further steps to obtain a tractable lower-bound \cite{celik2022specializing} 
\begin{align}\label{eq::LBComps_with_aux}
    \max_{\expert{\pi}} & \mathbb{E}_{\contextcomp{\pi},  \expert{\pi}} \left[\reward + \alpha \log \tilderesponsibility{\pi} \right] \nonumber \\
    &+\alpha\mathbb{E}_{\contextcomp{\pi}}\left[\textrm{H}\left[\expert{\pi}\right]\right]
\end{align}
for the expert $\expert{\pi}$ updates and a lower-bound for the per-expert context $\contextcomp{\pi}$ updates
\begin{align}\label{eq::LBCtxtDistr_with_aux}
    \max_{\contextcomp{\pi}}&  \mathbb{E}_{\contextcomp{\pi}} \left[L_c(\option,\context) + (\beta - \alpha) \log \tildegating{\pi}\right]  \nonumber\\
    &+\beta \textrm{H}\left(\contextcomp{\pi}\right),
\end{align}
where  $ L_c(\option, \context) = \mathbb{E}_{\expert{\pi}} \big[\reward + \alpha \log \tilderesponsibility{\pi}\big] + \alpha \textrm{H}\left[\expert{\pi}\right]$. The variational distributions $\tilderesponsibility{\pi}= \pi_{old}(\option|\context, \mpparam)$ and $\tildegating{\pi} = \pi_{old}(\option|\context) $ arise through the decomposition and are responsible for learning diverse solutions and concentrating on context regions with small, or no support by $\contextdistr{\pi}$ \cite{celik2022specializing}. In every iteration, the variational distributions are updated in closed form to tighten the bounds. Details of the equations are in the Appendix \ref{sec::Appendix::Self-Paced_Sill_Learning}.

\section{Related Work}
\textbf{Contextual Episode-based Policy Search (CEPS).} CEPS is a black-box approach to reinforcement learning (RL), in which the search distribution is the agent's policy that maps the contexts to controller parameters, typically represented as motion primitives \citep{schaal2006dynamic, paraschos2013probabilistic, li2023prodmp}. One of the noteworthy advantages of CEPS lies in the independence of assumptions such as the Markovian property in common MDPs. This characteristic renders it a versatile methodology, particularly well-suited for addressing a diverse array of intricate tasks where the formulation of a Markovian reward function is difficult \cite{otto2023deep}. CEPS has been explored by applying various optimization techniques, including Policy Gradients \citep{sehnke2010parameter}, Natural Gradients \citep{wierstra2014natural}, stochastic search strategies \citep{hansen2001completely, mannor2003cross, abdolmaleki2019contextual}, and trust-region optimization techniques \citep{abdolmaleki2015model, daniel2012hierarchical, tangkaratt2017policy}, particularly in the non-contextual setting.
Researchers extended the setting by incorporating linear \citep{tangkaratt2017policy, abdolmaleki2019contextual} and non-linear contextual adaptation \citep{otto2023deep, li2023open}, leveraging the recently introduced trust-region layers for neural networks \citep{otto2021differentiable}. The work by \citep{li2023open} additionally introduces step-wise updates to improve sample-efficiency. However, all previously mentioned methods learn single-mode policies and do not address acquiring diverse skills leveraging automatic curriculum learning.

\textbf{Curriculum Reinforcement Learning.} Curriculum reinforcement learning can potentially increase the performance of RL agents, especially in sparse-rewarded environments \cite{tao2024rfcl} in which exploration is fundamentally difficult. Adapting the environment based on the agent's learning process has been proposed by several works already, e.g. automatically generating sets of tasks or goals to increase the learning speed of the agent \citep{florensa2017reverse, florensa2018automatic,sukhbaatar2018intrinsic, zhang2020automatic,  wohlke2020performance, Racaniere2020Automated}, or generating a curriculum by interpolating an auxiliary and known distribution of target tasks \citep{klink2022curriculum, klink2020self, klink2020selfdeep, Klink2024TransCurr}. Works propose sampling a training level from a prespecified set of environments \citep{jiang2021prioritized}, or unsupervised environment design \citep{jiang2021replay, dennis2020emergent} based on the agent's learning process. The work by \citep{klink2021boosted} proposes improving the approximation of the state-action value function by representing it as a sum of residuals acquired in previous curriculum tasks. None of the above methods apply automatic curriculum learning on an RL problem with an MoE policy, except for the work in \citep{celik2022specializing}. However, they parameterize the curriculum distribution as Gaussian, suffering from low representation capacity and requiring knowledge about the environment's context distribution. Instead, we leverage energy-based models to avoid these shortcomings. 

\textbf{RL with Mixture of Experts (MoE).}\citet{ren2021probabilistic} propose using MoE policy representation and presents a novel gradient estimator to calculate the gradients w.r.t. the MoE parameters. \citet{huang2023reparameterized} present a model-based RL approach to train latent variable models. The work presents a novel lower bound for training the multi-modal policy parameterization. Recently, \citet{hendawy2023multi} proposed using MoEs for learning a shared representation in multi-task reinforcement learning, whereas \citet{akrour2021continuous} present how interpretable MoEs can be learned in continuous RL. These methods differ from our work in that they are not categorized in the CEPS framework, or are model-based variants and do not use automatic curriculum learning techniques. In the CEPS framework, RL with MoE policies has also been explored in the works by \citet{daniel2012hierarchical, end2017layered}, in which an MoE model with linear experts without automatic curriculum learning is learned. Additional constraints need to be added to enforce diversity in the experts. In the work by \citet{tosatto2021contextual} a mixture model is used to perform RL, however, pre-recorded demonstration data is required to train the mixture model and no curriculum learning is considered. Related method to MoEs, Product of Experts was used in \cite{hansel2023hierarchical, le2023hierarchical} for motion generation.\\
The work by \citet{celik2022specializing} also uses MoE policies and relies on the maximum entropy objective as we do, however, their method only considers linear experts with Gaussian per-expert distributions which limits the performance and consequently requires many experts to solve a task. Moreover, it requires environment knowledge to hand-tune a punishment term to keep the optimization of the per-expert context distributions within the context bounds.

\textbf{Quality-Diversity Optimization (QDO).} Learning diverse skills has also been explored in the evolutionary strategy community, most notably with the MAP-Elites algorithm \cite{cully2015robots}, where behavioral descriptors are defined to distinguish the different learned motions. Extensions \cite{nilsson2021policy, faldor2023map, faldor2023synergizing} have been proposed to improve the performance of these methods. However, these methods can not easily be applied to the contextual setting where different controller parameters should be chosen in different situations such that post hoc adaptations \cite{keller2020model, faldor2023synergizing} are required. In contrast to QDO methods, in our work diversity measurement naturally arises through the considered objective and does not need defining behavioral descriptors. Moreover, Di-SkilL indirectly learns a gating distribution that selects the expert after observing a context. 

\textbf{Unsupervised Reinforcement Learning (URL).} URL also considers learning diverse policies \citep{yang2024task, eysenbach2021information, urlb, eysenbach2018diversity, CamposTXSGT20, lee2019efficient, aps-liu21b}. The objective differs from ours and skills are trained in the absence of an extrinsic reward. We discuss parallels in the Appendix \ref{sec::Appendix::RelatedWork}. 

\section{Diverse Skill Learning}\label{sec::VSL}
The common Contextual Episodic Policy Search (CEPS) loop \citep{kupcsik2013data} with a Mixture of Experts (MoE) policy representation learning observes a context $\context$, and then selects an expert $\option$ that subsequently adjusts the controller parameters $\mpparam$ given ($\context$, $\option$). We consider the same process during testing time, as shown in blue color in Fig. \ref{fig::overview} (see also Fig. \ref{fig::pgm_inference}). 
However, the procedure changes during training for Di-SkilL as automatic curriculum learning requires that the agent can determine which context regions it prefers to focus on.
In this case, we observe a batch of context samples from the environment's context distribution $\contextdistr{p}$. For each of these samples, every per-expert context distribution $\contextcomp{\pi}$ calculates a probability, which results in a categorical distribution over the contexts $\context$. We use these probabilities to sample contexts $\context_T$ for the corresponding expert $\option$ resulting in ($\context_T, \option$) sample pairs (see orange parts in Fig. \ref{fig::overview} and Fig. \ref{fig::pgm_training}).
Each chosen expert $\option$ provides Gaussian distributions over the motion primitive parameters $\mpparam$ by mapping the contexts $\context_T$ to mean vectors $\cvec{\mu}_{\mpparam}$ and covariance matrices $\cvec{\Sigma}_{\mpparam}$ using a parameterized neural network. We can now sample motion primitive parameters $\mpparam$ from these Gaussian distributions to generate trajectories $\tau$ using a motion primitive generator. These trajectories are subsequently executed on the environment (green color in Fig. \ref{fig::overview}) and an episode return $R(\mpparam, \context_T)$ is observed and used for updating the MoE (see Section \ref{sec::VSL_UPDATE}).
Yet, there exist several issues for a stable overall training of the MoE model, which requires special treatment for each $\contextcomp{\pi}$ and $\expert{\pi}$. Algorithmic details and parameterizations of the model can be found in the Appendix \ref{sec::Appendix::additional_information}.

\begin{figure*}[t!]
   \begin{minipage}[b]{0.25\textwidth}
        \centering
       \resizebox{0.9\textwidth}{!}{\includegraphics[width=\linewidth]{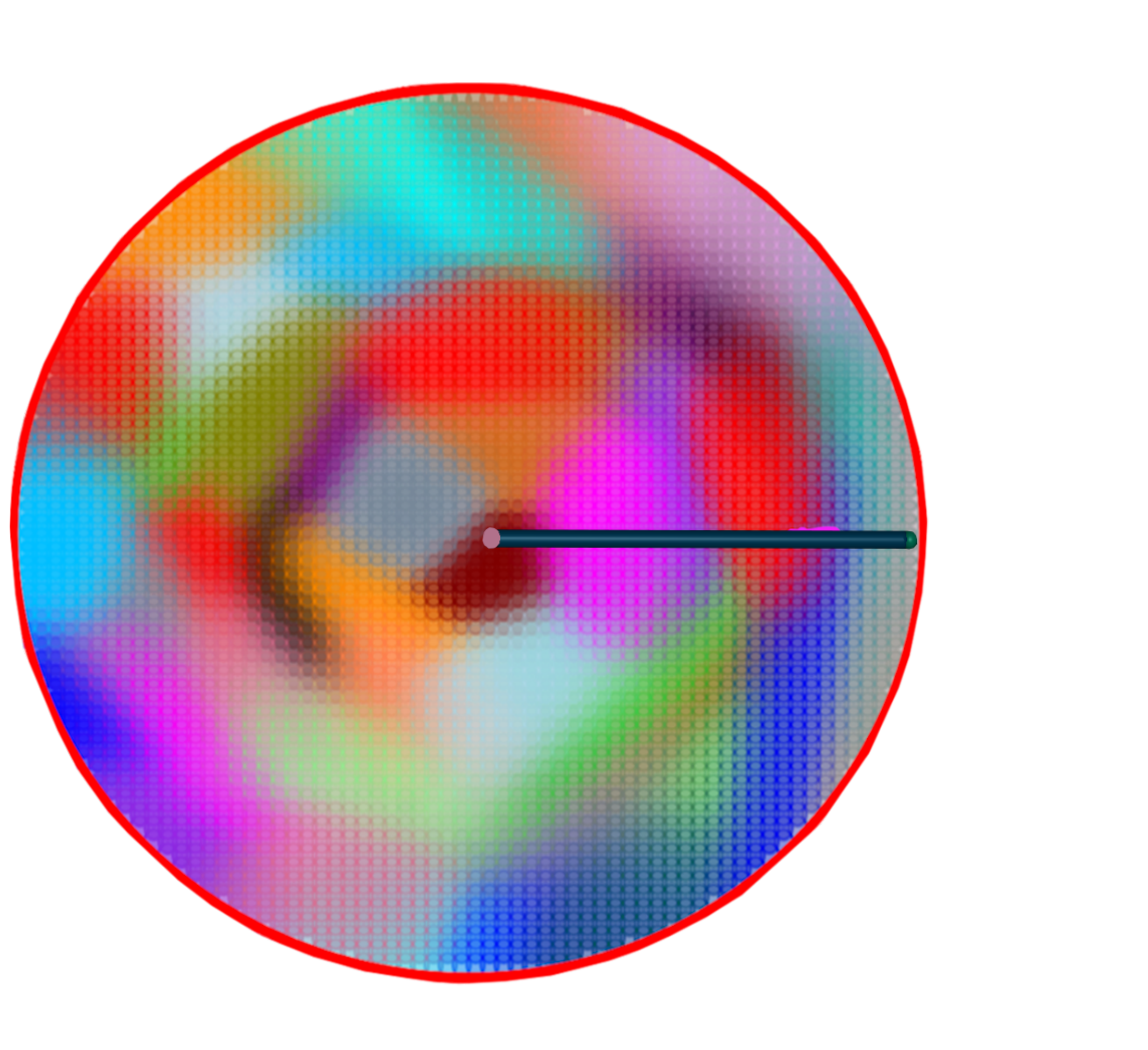}}
       \subcaption[]{}
       \label{fig::exps::analysis_ctxtdistrs}
   \end{minipage}\hfill
   \begin{minipage}[b]{0.25\textwidth}
        \centering
       \resizebox{0.9\textwidth}{!}{\includegraphics[width=\linewidth]{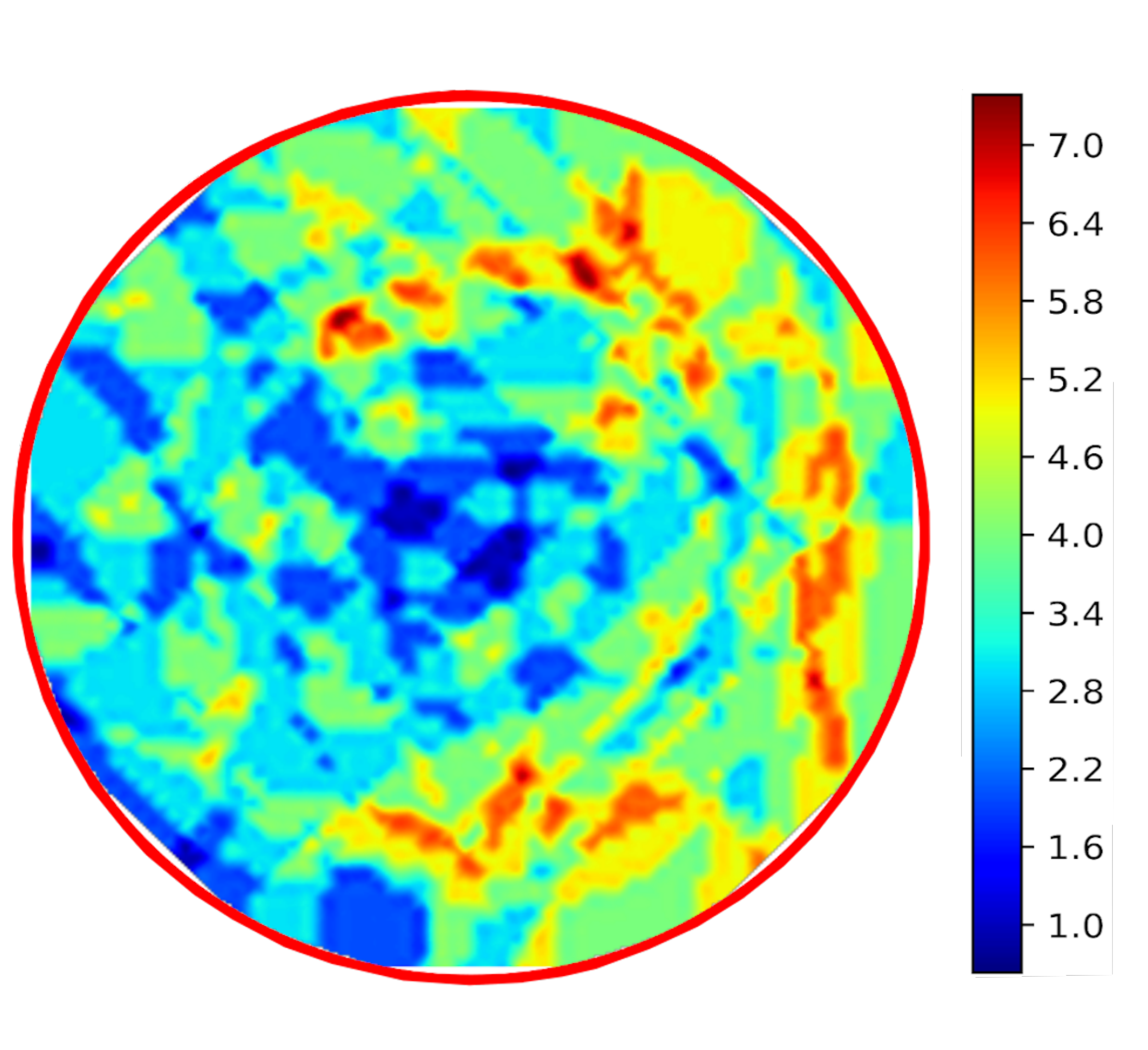}}
       \subcaption[]{}
       \label{fig::exps::analysis_frequency}
   \end{minipage}\hfill
   \begin{minipage}[b]{0.25\textwidth}
        \centering
       \resizebox{1\textwidth}{!}{\includegraphics[width=\linewidth]{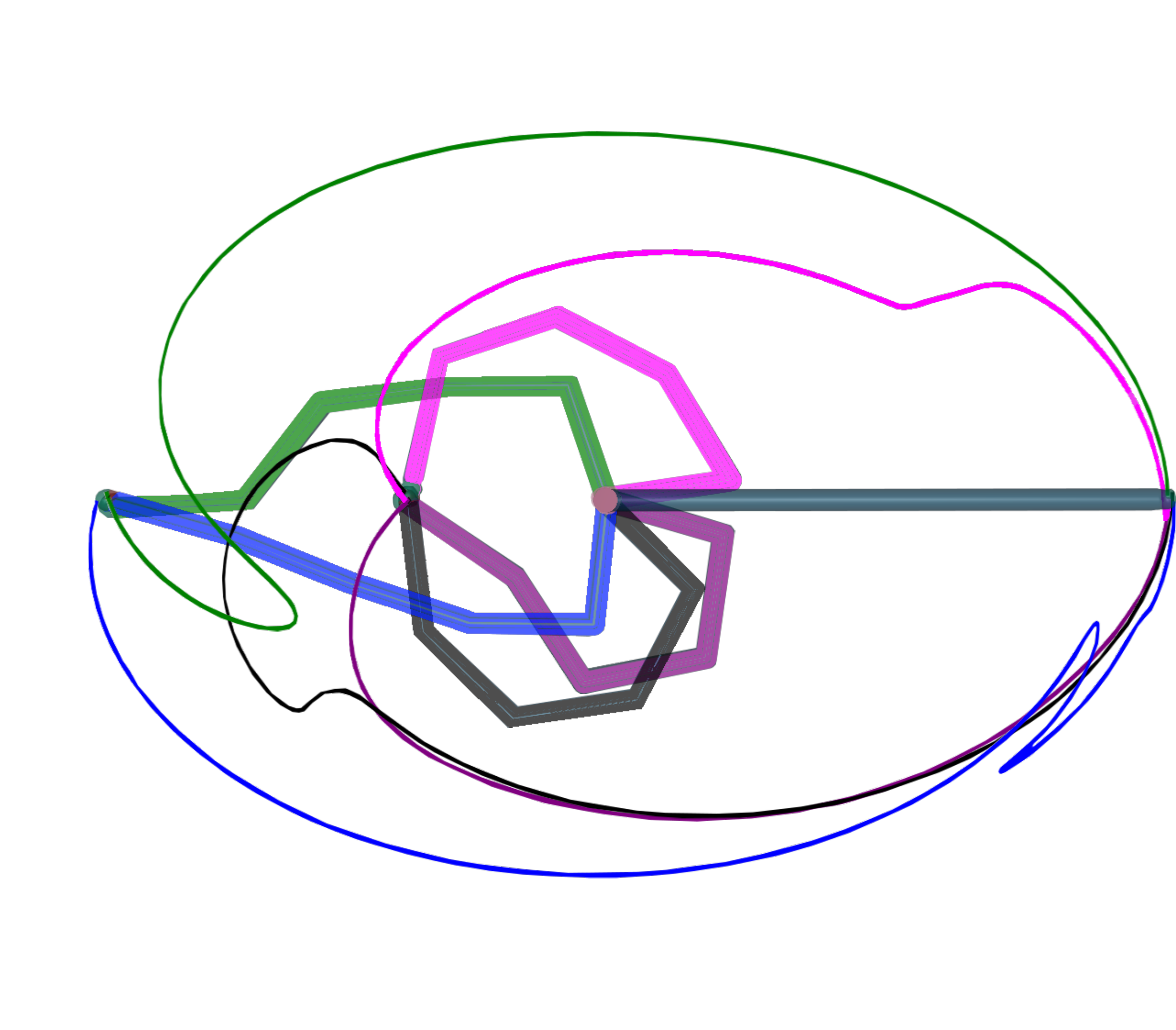}}
       \subcaption[]{}
       \label{fig::exps::diverse_tip_trajs}
   \end{minipage}\hfill
   \begin{minipage}[b]{0.25\textwidth}
        \centering
       \resizebox{0.9\textwidth}{!}{\input{figures/environments/5lr_env}}
       \subcaption[]{}
       \label{fig::exps::5lr}
   \end{minipage}\hfill
   \caption{\textbf{a)} High-probability regions of the individual per-expert context distributions $\contextcomp{\pi}$, where a color represents an expert $\option$. The red circle marks the context space of goal-reaching positions for the 5-Link Reacher's tip. The specialization of $\contextcomp{\pi}$ is induced by $\tildegating{\pi}$. \textbf{b)} Different $\contextcomp{\pi}$ need to overlap for learning diverse skills. This overlapping is induced by the entropy bonus $\textrm{H}\left[\contextcomp{\pi}\right]$. \textbf{c)} Different tip trajectories sampled in the same contexts. The trajectories and the end joint constellation are in the same color. The diversity in the parameter space is induced by $\tilderesponsibility{\pi}$. \textbf{d)} Visualization of the 5-Link Reacher task (\textbf{5LR}).}
   \label{fig::exps::diversity_analysis}
\end{figure*}

\subsection{Energy-Based Model For Automatic Curriculum Learning}
To illustrate these issues, we consider a bounded, uniformly distributed two-dimensional environment context distribution $p(\context)$ (see example in the Appendix \ref{sec::Appendix::additional_information} in Fig. \ref{fig::vsl::per_expert_context_distr_illustration}). It is challenging for a Reinforcement Learning (RL) agent to automatically learn its curriculum $\contextcomp{\pi}$ within the valid context space \cite{celik2022specializing}. Hard discontinuities such as steps often naturally arise in $\contextdistr{p}$ due to the environment's finite support in real-world environments. For instance, in an environment where the agent's task is to place an object in specific positions on a table, the probability of observing a goal position outside the table's surface is zero. 
This implies that a large subset of the context space has no probability mass. Therefore, exploration in these regions might be difficult if there is no guidance encoded in the reward.
Even if it is guaranteed that $\contextcomp{\pi}$ only samples valid contexts, it still needs to be able to represent multi-modal distributions, such as illustrated in Fig. \ref{fig::vsl::env_context_distr_illustration}. This multi-modality can be present because of environmental circumstances or simply if experts $\expert{\pi}$ prefer contexts in spatially apart regions. For the object placing example, this could correspond to regions on the table where the object cannot be placed due to obstacles or holes. 
We therefore require $\contextcomp{\pi}$ being able to represent \textbf{i)} complex distributions, \textbf{ii)} multi-modality and \textbf{iii)} only explore within the valid context bounds of $\contextdistr{p}$. 
We propose parameterizing each per-expert distribution $\contextcomp{\pi}$ as an energy-based model 
\begin{align}\label{eq:Energy-based-per-component-ctxt-distr}
    \contextcomp{\pi} = \exp(\potential{\phi}{\option})/ \normalizer
\end{align}
to address the issues i) and ii), where the energy function $\phi_o$ is a per-expert learnable neural network. Energy-based models (EBMs) have shown to be capable of representing sharp discontinued functions and multi-modal distributions \citep{florence2022implicit}. Yet, they are hard to train and sample from due to the intractable normalizing constant $\normalizer = \int_{\context} \exp(\potential{\phi}{\option}) d\context$. 
We can circumvent and address these issues iii) by approximating the normalizing constant with contexts $\context \sim \contextdistr{p}$ as $\normalizer \approx \sum_{i=1}^N \exp(\phi_{\option}(\context_i))$. This approximation is justified as we can sample from $\contextdistr{p}$ by simply resetting the environment without execution. Additionally, the EBM will encounter important parts of the context space during the training by resampling a large enough batch of contexts $\context \sim \contextdistr{p}$ in each iteration. Each expert can therefore sample preferred contexts from the current batch of valid contexts by calculating the probability for each of the contexts using $\contextcomp{\pi}$ as parameterized in Eq. \ref{eq:Energy-based-per-component-ctxt-distr}. Updating the parameters of the EBM can readily be addressed by the standard RL objective for diverse skill learning, as described in the next section. It should be noted that explicit models such as Gaussians, or Normalizing Flows \cite{papamakarios2021normalizing} can also be used to parameterize $\contextcomp{\pi}$, but their support cannot be easily restricted to a bounded space with hard discontinuities defined by the environment. Therefore, sampling from an explicit $\contextcomp{\pi}$ can easily generate invalid contexts, especially if the valid distribution has hard non-linearities. 

\subsection{Updating the Mixture of Experts Model}\label{sec::VSL_UPDATE}
We update each expert $\expert{\pi}$ and its corresponding per-expert context distribution $\contextcomp{\pi}$ by maximizing the objectives in Eq. \ref{eq::LBComps_with_aux} and in Eq. \ref{eq::LBCtxtDistr_with_aux}, respectively. These decomposed objectives allow us to independently update both distributions and to retain the properties of diverse skill learning from the objective in Eq. \ref{eq::maxEntrObj}. 
However, updating the distributions is not straightforward due to the bi-level optimization that leads to a dependency on both terms. This is particularly challenging for the expert $\expert{\pi}$ as the sampled contexts $\context$ can drastically change from one iteration to another if $\contextcomp{\pi}$ changes too aggressively. The same applies for updating $\contextcomp{\pi}$ as calculating the objective requires calculating an integral over $\mpparam$ under the expectation of $\expert{\pi}$.
For a stable update for both distributions, we employ trust-region updates to restrict the change of both distributions from one iteration to another. These updates have been shown to improve the learning process \citep{peters2010relative,  schulman2015trust, schulman2017proximal, otto2021differentiable}. 

\textbf{Expert Update.} We parameterize each expert $\expert{\pi}$ with a single neural network and update them by a trust-region constrained optimization 
\begin{align}\label{eq::LBComps_with_aux_with_TRConstraint}
    \max_{\expert{\pi}} & \mathbb{E}_{\contextcomp{\pi},  \expert{\pi}} \left[\reward + \alpha \log \tilderesponsibility{\pi} \right] \nonumber \\
    &\quad \quad \quad \quad \quad  +\alpha\mathbb{E}_{\contextcomp{\pi}}\left[\textrm{H}\left[\expert{\pi}\right]\right] \\
    & \st \quad  \KL{\expert{\pi}}{\old{\pi}(\mpparam|\context, \option)} \le \epsilon ~~~~~ \forall ~\context \in \mathcal{C} \nonumber,
\end{align}
where the KL-bound ensures that the expert $\expert{\pi}$ does not differ too much from the expert $\old{\pi}(\mpparam|\context, \option)$ from the iteration before for each context $\context$. The entropy bonus $\textrm{H}\left[\expert{\pi}\right]$ incentivizes $\expert{\pi}$ to fully cover the parameter space while avoiding $(\mpparam, \context)$ regions that are covered by other experts $\option$. The latter is guaranteed by $\tilderesponsibility{\pi}$ which rewards $(\mpparam, \context)$ regions that can be assigned to expert $\option$ with high probability. We efficiently update the experts using trust region layers \cite{otto2021differentiable, otto2023deep}. 

\textbf{Per-Expert Context Distribution Update.} We consider the objective with the augmented rewards as shown in Eq. \ref{eq::LBCtxtDistr_with_aux} for updating each context distribution $\contextcomp{\pi}$. We can not apply the trust region layers \citep{otto2021differentiable} in this case, as $\contextcomp{\pi}$ is a discrete distribution over the context samples $\mathbf c_i$ parameterized by the EBM. Yet, we can still use PPO \citep{schulman2017proximal} for updating $\contextcomp{\pi}$ and simplifying our objective, as we can now calculate many terms in closed form. For this, we rewrite the objective as 
\begin{align}\label{eq::LBCtxtDistr_with_aux_EBM}
    \max_{\contextcomp{\pi}}&  \sum_{\mathbf c_i \sim p(\mathbf c)} \pi(\mathbf{c}_i|o) %
    L_c(\option,\context_i) + \\
    &\sum_{\mathbf c_i \sim p(\mathbf c)} \pi(\mathbf{c}_i|o)\big( (\beta - \alpha)  \log \tilde{\pi}(o|\mathbf{c}_i) 
      - \beta \log \pi(\mathbf{c}_i|o)\big) \nonumber
\end{align}
and observe that all terms in the second sum can be calculated in closed form. Note that the first term is approximated by resampling the context samples using  $\contextcomp{\pi}$ since computing $L_c(\option, \context)$ requires calculating the integral over $\mpparam$ under the expectation of $\expert{\pi}$ as 
$ L_c(\option, \context) = \mathbb{E}_{\expert{\pi}} \big[\reward   + \alpha \log \tildegating{\pi}\big] + \alpha \textrm{H}\left[\expert{\pi}\right]$. This expectation can only be estimated for context vectors that are actually chosen by the component. 
The entropy bonus in Eq. (\ref{eq::LBCtxtDistr_with_aux_EBM}) incentivizes covering of the context space, while focusing on context regions that are not, or only partly covered by other options. The latter is guaranteed by $\tildegating{\pi}$ which assigns a high probability if expert $\option$ can be assigned to the context $\context$.

\begin{figure*}[t!]
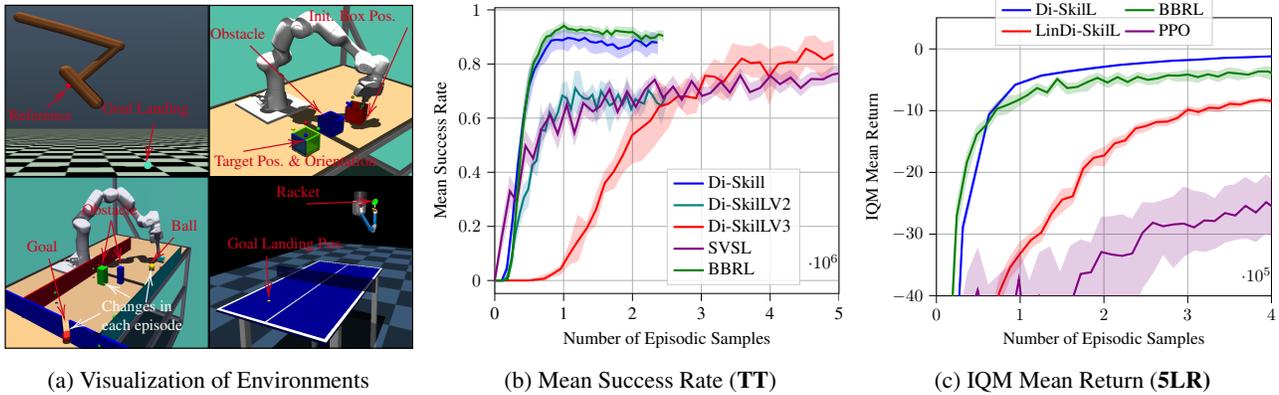

   \begin{minipage}[b]{0.33\textwidth}
        \centering
       \resizebox{\textwidth}{!}{\input{figures/environments/all_envs_icml}}
       \subcaption[]{Visualization of Environments}
       \label{fig::exps::envs}
   \end{minipage}\hfill
   \begin{minipage}[b]{0.33\textwidth}
        \centering
       \resizebox{\textwidth}{!}{\input{figures/experiments/ablations/curriculum_ablations_tt}}
       \subcaption[]{Mean Success Rate (\textbf{TT})}
       \label{fig::exps_ablation::success_rate}
   \end{minipage}\hfill
   \begin{minipage}[b]{0.33\textwidth}
        \centering
       \resizebox{\textwidth}{!}{\input{figures/experiments/reacher/rewards}}
       \subcaption[]{IQM Mean Return (\textbf{5LR)}}
       \label{fig::exps::rewards_5lre}
   \end{minipage}\hfill
   \caption{\textbf{a)} (left top) Hopper Jump Task (\textbf{HJ}). (Top right) Box Pushing with Obstacle (\textbf{BPO}). (Bottom Left) Robot Mini Golf (\textbf{MG}). (Bottom right) Robot table tennis (\textbf{TT}).
   \textbf{b)} Ablation studies, showcasing the need for automatic curriculum learning for Di-SkilL. BBRL and Di-SkilL can solve TT environment decently. Di-SkilL's variants without curriculum learning struggle to achieve a good performance. SVSL needs more samples to achieve around 80\% success rate, suffering under the linear experts. \textbf{c)} Performance of Di-SkilL, BBRL, LinDi-SkilL, and PPO on 5LR with sparse in-time rewards. }
\end{figure*}

\subsection{How does Diversity Emerge?}\label{sec::Diversity_Emerge}
From the Eq. \ref{eq::LBComps_with_aux_with_TRConstraint} and Eq. \ref{eq::LBCtxtDistr_with_aux_EBM} it is clear that diverse behaviors, represented by the experts, emerge from the interplay of those terms in Eq. \ref{eq::LBComps_with_aux_with_TRConstraint} and Eq. \ref{eq::LBCtxtDistr_with_aux_EBM}. We visually demonstrate the meaning of the individual terms on the 5-Link Reacher task (see Fig. \ref{fig::exps::5lr}). The Reacher needs to reach a goal position in the two-dimensional space with its tip. In this task, a context represents the goal position within the context space, visualized as a red circle around the reacher's fixed first joint (Fig. \ref{fig::exps::analysis_ctxtdistrs}). We trained Di-SkilL with 50 experts.    

In Fig. \ref{fig::exps::analysis_ctxtdistrs} we show the high-probability regions of the individual per-expert context distributions $\contextcomp{\pi}$, by setting the color intensity proportional to this probability. Each color represents an individual expert $\option$. 
Each $\contextcomp{\pi}$ concentrates on a sub-region of the context space such that the corresponding $\expert{\pi}$ becomes an expert there. This specialization is incentivized by the term $\tildegating{\pi}$ in Eq. \ref{eq::LBCtxtDistr_with_aux_EBM}. However, for learning diverse behaviors for the same context regions, it is necessary that the per-expert context distributions $\contextcomp{\pi}$ overlap, which is motivated by the entropy term $\textrm{H}\left[\contextcomp{\pi}\right]$ in Eq. \ref{eq::LBCtxtDistr_with_aux_EBM}.

These overlapping context regions are visualized in Fig. \ref{fig::exps::analysis_frequency}, where we count how many experts $\option$ are active for each context. The figure shows that more experts prefer regions close to the initial position of the reacher, indicating that these contexts are easier to solve. Despite the closeness to the reacher's initial position, the agent does not have to exert much energy to reach these points. Indeed, both aspects are present in the task's reward function (see Appendix \ref{sec::Appendix:ExpDetails} for details), explaining why the left half plane of the context space has fewer overlapping. However, the learned MoE has two or more experts active in most parts of the context region. These experts differ in their behavior (see Fig. \ref{fig::exps::diversity_analysis} for examples), which is motivated by the terms $\tilderesponsibility{\pi}$ and $\textrm{H}\left[\expert{\pi}\right]$ in Eq. \ref{eq::LBComps_with_aux_with_TRConstraint}.

\section{Experiments}
In our evaluations, we compare Di-SkilL against the baselines \textbf{BBRL} \citep{otto2023deep} and \textbf{SVSL} \citep{celik2022specializing}. Whenever the environment satisfies the Markov properties, we additionally compare against \textbf{PPO} \cite{schulman2017proximal}. BBRL and SVSL are suitable baselines as they are state-of-the-art CEPS algorithms that can learn complex skills. BBRL can learn highly non-linear policies leveraging trust region updates. SVSL learns a linear Mixture of Experts (MoE) model and can capture multi-modality in the behavior space. We consider challenging robotic environments with continuous context and parameter spaces. The considered environments either have a non-Markovian reward function, i.e. require retrospective data for calculation, or temporally sparse reward functions, increasing the learning complexity due to more difficult exploration. 

We start by providing an overview of the benchmarking environments. In an ablation study, we then show that automatic curriculum learning is an essential feature of Di-SkilL to learn high-performing skills. 
Lastly, on five sophisticated robot simulation tasks we report the \textbf{performance} of Di-SkilL against the baselines. The results show that Di-SkilL performs on par, or better than the baselines on all tasks. In addition to the performance analysis, we qualitatively show Di-SkilL's learned \textbf{diverse skills} on the challenging table tennis, box pushing, and reaching tasks.

\subsection{Environments}
The considered environments are visualizations in Fig. \ref{fig::exps::envs}.  Throughout all environments, we used ProDMPs \cite{li2023prodmp} to generate trajectories (see Appendix \ref{sec::Appendix::additional_information}). Detail descriptions are provided in the Appendix \ref{sec::Appendix:ExpDetails}.

\textbf{Table Tennis (TT).} A 7-degree of freedom (DoF) robot has to learn fast and precise motions to hit the ball to a desired position on the opponent's side. The 4-dim. context consists of the incoming ball's landing position and the desired ball's landing position on the opponent's side. The TT env. requires good exploratory behavior and has a non-Markovian reward structure, making step-based approaches infeasible to learn useful skills \citep{otto2023deep}.

\textbf{Table Tennis Hard (TT-H).} We extend the TT environment to a more challenging version by varying the ball's initial velocity. This additionally increases the learning complexity, as the agent now needs to reason about the physical effects of changed velocity ranges and requires improved adaptability.

\textbf{5-Link Reacher (5LR).} The 5-Link reacher has to reach a goal position within all quadrants in the context space (see Fig. \ref{fig::exps::analysis_ctxtdistrs}) as opposed to the version in \cite{otto2023deep}, where the multi-modality in the behavior space (see Fig. \ref{fig::exps::diverse_tip_trajs}) was avoided by constraining the context space to the upper half of the context space. Additionally, the time-sparse reward makes this task a challenging benchmark.

\textbf{Hopper Jump (HJ).} Presented in \cite{otto2023deep} in which the Hopper \cite{gym} is tasked to jump as high as possible while landing in a goal position. The HJ has a non-Markovian reward, making step-based RL methods unfeasible to learn useful policies \citep{otto2023deep}.

\textbf{Box Pushing with Obstacle (BPO).} A 7DoF robot is tasked to push a box to a target position and rotation while avoiding an obstacle. In addition to the time-spare reward \cite{otto2023deep}, our version includes the obstacle and considers a larger range of the box's target position. 

\textbf{Robot Mini Golf (MG).} The 7DoF robot has to hit the ball in an env. with two obstacles (static, varying), such that it passes the tight goal. The context is the obstacle's, the goal's, and the ball's position. The MG environment has a non-markovian reward, making step-based RL methods unfeasible to learn useful policies \citep{otto2023deep}.

\begin{figure*}[t!]
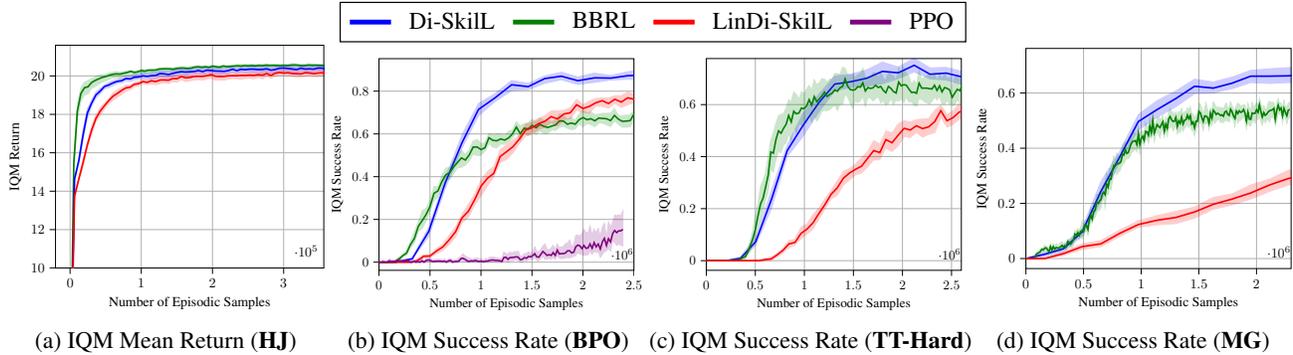

    \centering
    \resizebox{0.5\textwidth}{!}{
    \definecolor{ao(english)}{rgb}{0.0,0.5,0}
\begin{tikzpicture} 
    \begin{axis}[%
    hide axis,
    xmin=10,
    xmax=50,
    ymin=0,
    ymax=0.4,
    legend style={
        draw=white!15!black,
        legend cell align=left,
        legend columns=-1, 
        legend style={
            draw=black,
            column sep=1ex,
            line width=0.5pt
        }
    },
    ]
    \addlegendimage{line width=2pt, color=blue}
    \addlegendentry{Di-SkilL};
    \addlegendimage{line width=2pt, color=ao(english)}
    \addlegendentry{BBRL};
    \addlegendimage{line width=2pt, color=red}
    \addlegendentry{LinDi-SkilL};
    \addlegendimage{line width=2pt, color=violet}
    \addlegendentry{PPO};
    \end{axis}
\end{tikzpicture}}%

    \begin{minipage}[b]{0.25\textwidth}
        \centering
       \resizebox{1\textwidth}{!}{\input{figures/experiments/hj/mean_rewards}}
       \subcaption[]{IQM Mean Return ({\bf HJ})}
       \label{fig::exps::HJ_mean_rewards}
    \end{minipage}\hfill
   \begin{minipage}[b]{0.25\textwidth}
        \centering
       \resizebox{1\textwidth}{!}{\input{figures/experiments/bp_obs/success_rates}}
       \subcaption[]{IQM Success Rate ({\bf BPO})}
       \label{fig::exps::BPOBS_success_rate}
    \end{minipage}\hfill
    \begin{minipage}[b]{0.25\textwidth}
        \centering
       \resizebox{0.97\textwidth}{!}{\input{figures/experiments/tt_VEL/success_rates}}
       \subcaption[]{IQM Success Rate ({\bf TT-Hard})}
       \label{fig::exps::ttVel_success_rate}
   \end{minipage}\hfill
   \begin{minipage}[b]{0.25\textwidth}
        \centering
       \resizebox{1\textwidth}{!}{\input{figures/experiments/mg/success_rates}}
       \subcaption[]{IQM Success Rate ({\bf MG})}
       \label{fig::exps::MG_success_rate}
   \end{minipage}\hfill
   \caption{\textbf{Performance on the a)} \textbf{HJ} (Hopper Jump)  \textbf{b)} \textbf{BPO} (Box Pushing with Obstacle), \textbf{c)} \textbf{TT-H} (Table Tennis Hard), and \textbf{d)} \textbf{MG} (Robot Mini Golf) tasks. \textbf{a)} Di-SkilL performs on par with BBRL on the HJ task. \textbf{b)} The multi-modality introduced by the obstacle in the box pushing task leads to worse performance for BBRL than for Di-SkilL and LinDi-SkilL. PPO suffers under the time-sparse reward setting. \textbf{c)} While BBRL converges faster, Di-SkilL achieves a higher success rate eventually. \textbf{d)} Di-SkilL outperforms the baselines on the \textbf{MG} task. LinDi-SkilL performs poorly on the non-Markovian rewarded tasks TT-H and MG, indicating that highly non-linear policies are beneficial.
   }
   \label{fig::exps::TTBPIQM_TTSTRIKES}
\end{figure*}

\subsection{ACL Benefits}
Automatic Curriculum learning (ACL) enables Di-SkilL's experts to shape their curriculum by explicitly sampling from preferred context regions. We analyze the importance of this feature by comparing the performance of variants of Di-SkilL on the table tennis (TT) task. 

For both variants \textit{Di-SkilLV2} and \textit{Di-SkilLV3} we disable ACL by fixing the term induced by the variational distribution to $\log\tildegating{\pi}=0$ in Eq. \ref{eq::LBCtxtDistr_with_aux_EBM} and by setting the entropy scaling parameter $\beta=2000$.
Ignoring the variational distribution $\tildegating{\pi}$ during training eliminates the intrinsic motivation of the per-expert context distribution $\contextcomp{\pi}$ to focus on sub-regions in the context space that are not, or only partially, covered by any other $\contextcomp{\pi}$ (Section \ref{sec::Diversity_Emerge}). Setting $\beta=2000$ incentivizes each $\contextcomp{\pi}$ to maximize its entropy, resulting in a uniform distribution in the environment's bounded context space. 
For Di-SkilL we keep the ACL and set $\beta=4$. We provide the same number of 50 context-parameter samples per expert for \textit{Di-SkilLV2} and Di-SkilL, whereas \textit{Di-SkilLV3} receives 260 samples per expert in each iteration. All variants possess 5 experts. \\
In Fig. \ref{fig::exps_ablation::success_rate} we report the mean success rates and the 95\% confidence interval for each method on at least 4 seeds. \textit{Di-SkilLV2} converges to a much smaller success rate, and \textit{Di-SkilLV3} needs more samples to reach the level of Di-SkilL. BBRL and Di-SkilL achieve high success rates, while BBRL performs slightly better. SVSL shows worse performance, even though the model has 20 experts. The results indicate that ACL is an essential feature of Di-SkilL ensuring that Di-SkilL can learn high-perfroming skills with fewer samples. 
Moreover, SVSL's poor performance shows that Gaussian parameterized per-expert context distributions that require additionally tuned punishment terms for guided updates are together with linear experts incapable of achieving a satisfying performance. 
\begin{figure*}[t!]
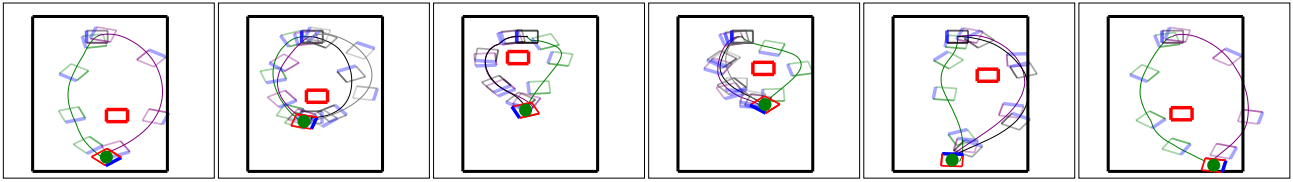

    \begin{minipage}[b]{0.166\textwidth}
        \centering
       \resizebox{\textwidth}{!}{\input{figures/experiments/bp_obs/box_positions/box_trajs_1}}
   \end{minipage}\hfill
   \begin{minipage}[b]{0.166\textwidth}
        \centering
       \resizebox{\textwidth}{!}{\input{figures/experiments/bp_obs/box_positions/box_trajs_2}}
   \end{minipage}\hfill
   \begin{minipage}[b]{0.166\textwidth}
        \centering
       \resizebox{\textwidth}{!}{\input{figures/experiments/bp_obs/box_positions/box_trajs_3}}
   \end{minipage}\hfill
   \begin{minipage}[b]{0.166\textwidth}
        \centering
       \resizebox{\textwidth}{!}{\input{figures/experiments/bp_obs/box_positions/box_trajs_4}}
   \end{minipage}\hfill
   \begin{minipage}[b]{0.166\textwidth}
        \centering
       \resizebox{\textwidth}{!}{\input{figures/experiments/bp_obs/box_positions/box_trajs_5}}
   \end{minipage}\hfill
   \begin{minipage}[b]{0.166\textwidth}
        \centering
       \resizebox{\textwidth}{!}{\input{figures/experiments/bp_obs/box_positions/box_trajs_6}}
   \end{minipage}\hfill  
    \caption{Di-SkilL's \textbf{Diverse Skills} for the Box Pushing with Obstacle \textbf{BPO} Task. The figures visualize diverse solutions to the same contexts $\context$ on a table (black rectangle). The red, thick rectangle represents the obstacle. The 7DoF robot is tasked to push the box (shown in different colors for each solution found) to the goal box position (red rectangle with a green dot) and align the blue edges to match the orientation. The context consists of the 2-dim. obstacle position, the 2-dim. goal position and the 1-dim. goal orientation around the z-axis. We visualized successful box trajectories for each sampled skill from the same Di-SkilL policy with 10 experts. The diversity learned in the parameter space results in different box trajectories ranging in position and orientation.
    }
       \label{fig::exps::diverse_box_trajectories}
\end{figure*}
\begin{figure*}[t!]
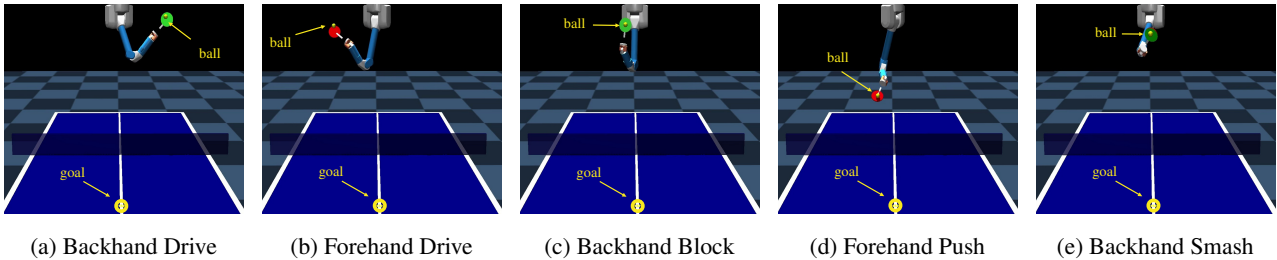

   \begin{minipage}[b]{0.2\textwidth}
        \centering
       \resizebox{\textwidth}{!}{\input{figures/experiments/tt_strikes_icml/strike_1}}
       \subcaption[]{Backhand Drive}
       \label{}
    \end{minipage}\hfill
   \begin{minipage}[b]{0.2\textwidth}
        \centering
       \resizebox{\textwidth}{!}{\input{figures/experiments/tt_strikes_icml/strike_2}}
       \subcaption[]{Forehand Drive}
       \label{}
    \end{minipage}\hfill
    \begin{minipage}[b]{0.2\textwidth}
        \centering
       \resizebox{\textwidth}{!}{\input{figures/experiments/tt_strikes_icml/strike_3}}
       \subcaption[]{Backhand Block}
       \label{}
   \end{minipage}\hfill
   \begin{minipage}[b]{0.2\textwidth}
        \centering
       \resizebox{\textwidth}{!}{\input{figures/experiments/tt_strikes_icml/strike_4}}
       \subcaption[]{Forehand Push}
       \label{}
   \end{minipage}\hfill
   \begin{minipage}[b]{0.2\textwidth}
        \centering
       \resizebox{\textwidth}{!}{\input{figures/experiments/tt_strikes_icml/strike_5}}
       \subcaption[]{Backhand Smash}
       \label{}
   \end{minipage}\hfill
   \caption{Di-SkilL's \textbf{Diverse Skills} for the Table Tennis Hard \textbf{TT-H} task. We fixed the ball's desired landing position and varied the serving landing position and the ball's initial velocity. Di-SkilL can return the ball in different striking types such as backhand or forehand strikes, where hitting the ball with the green side of the racket is referred to as backhand and forehand otherwise. The shown striking styles are captured from the same Di-SkilL policy that was trained with 10 experts. 
   }
   \label{fig::exps::diverse_strikes}
\end{figure*}

\subsection{Analyzing the Performance and Diversity}
For a detailed analysis, we have evaluated all methods on \textit{24 seeds} for each environment and algorithm and report the \textit{interquantile mean} (IQM) with a 95\% stratified bootstrap confidence interval as suggested by \cite{agarwal2021deep}. Please note that SVSL requires designing a punishment function to guide the context samples in the environment's valid context region, which makes its application difficult, especially if the context influences the objects' physics. We therefore propose comparing against LinDi-SkilL instead of SVSL. LinDi-SkilL also has linear experts but benefits from Di-SkilL's energy-based per-expert context distribution $\contextcomp{\pi}$ eliminating the need for punishment functions. 

The performance curve of the \textbf{HJ} task in Fig. \ref{fig::exps::HJ_mean_rewards} shows that Di-SkilL performs on par with BBRL, while BBRL converges slightly faster. Both methods can solve the task, indicating that the task doesn't require diversity. We can also see that LinDi-SkilL achieves a similar performance as BBRL and Di-SkilL, but needs more samples to converge. We provide additional analysis of this task in Appendix \ref{sec:appendix:additional_evaluations}.   

Fig. \ref{fig::exps::BPOBS_success_rate} shows the performance curves on the \textbf{BPO} task. The obstacle introduces multi-modality in the behavior space which cannot be captured by a single-mode policy. This multi-modality explains why DiSkilL and LinDi-SkilL outperform BBRL, while Di-SkilL still achieves the highest success rate. PPO's poor performance indicates that time-correlated exploration as used with motion primitives is effective in sparse rewarded tasks. 

A similar performance behavior can be observed in the \textbf{5LR} task. In Fig. \ref{fig::exps::rewards_5lre} we report the achieved returns and observe that Di-SkilL outperforms BBRL due to the ability to capture multi-modal behaviors (e.g. reaching from different sides) while PPO suffers from the sparse rewarded setting. Moreover, LinDi-SkilL's linear experts cause slow convergence, indicating that more experts are needed to effectively cover the whole context space. 
For both tasks, Di-SkilL's diverse skills in the parameter space $\mpparam$ induce different behaviors. Fig. \ref{fig::exps::diverse_box_trajectories} shows diverse box trajectories to several fixed goal and obstacle positions in the BPO task, whereas Fig. \ref{fig::exps::diverse_tip_trajs} shows different tip trajectories to several fixed goal positions in the 5LR task. 

The non-Markovian rewarded tasks (\textbf{TT-H} and \textbf{MG}) show that non-linear policies as learned by BBRL and Di-SkilL are beneficial. Di-SkilL and BBRL perform similarly well on the TT-H task (see Fig. \ref{fig::exps::ttVel_success_rate}), where Di-SkilL achieves a slightly higher end success rate compared to BBRL. However, there is a clear performance gap between Di-SkilL and BBRL on the MG task (see Fig. \ref{fig::exps::MG_success_rate}) with Di-SkilL outperforming BBRL. In both tasks, LinDi-SkilL performs worse than Di-SkilL and BBRL indicating that linear experts are insufficient for solving these tasks. 

Di-SkilL can discover diverse striking styles in the table tennis task (TT-H). Fig. \ref{fig::exps::diverse_strikes} shows some of these learned skills.
Additional strike visualizations are in Appendix \ref{sec:appendix:additional_evaluations}.

\section{Conclusion and Future Work}
In this paper, we propose Diverse Skill Learning (Di-SkilL), a novel method for learning diverse skills using a contextual Mixture of Experts. Each expert automatically learns its curriculum by optimizing for a per-expert context distribution $\contextcomp{\pi}$. We have demonstrated challenges that arise through enabling automatic curriculum learning (ACR) and proposed parameterizing $\contextcomp{\pi}$ as energy-based models (EBMs) to address these challenges. Additionally, we provided a methodology to efficiently optimize these EBMs. We also proposed using trust-region updates for the deep experts to stabilize our bi-level optimization problem. In an ablation, we have shown that ACR is necessary for efficient and performant learning. Moreover, in sophisticated robot simulation environments, we have shown that our method can learn diverse skills while performing on par or better than the baselines. Currently, the major drawback of our approach is its inability to replan, causing failures in the tasks if the robot even has small collisions with objects. We intend to address this issue in future research. To improve the sample complexity of our approach, we additionally plan to use off-policy RL techniques. 

\section*{Acknowledgements}
The authors acknowledge support by the state of Baden-Württemberg through
bwHPC, as well as the HoreKa supercomputer funded by the Ministry of Science, Research and the Arts Baden-Württemberg and by the German Federal Ministry of Education and Research. This work has been supported by the DFG Collaborative Research Center 1574, Circular Factory for the Perpetual Product, and by the pilot program Core Informatics of the Helmholtz Association (HGF). 

\section*{Impact Statement}
This paper presents work whose goal is to advance the field of Machine Learning. There are many potential societal consequences of our work, none of which we feel must be specifically highlighted here.

\bibliography{references}
\bibliographystyle{icml2024}

\newpage
\appendix
\onecolumn
\clearpage

\begin{figure*}[t!]
 \begin{minipage}[b]{0.2\textwidth}
        \centering
        \resizebox{0.7\textwidth}{!}{
        \tikz{
         \node[latent] (mpparam) {$\mpparam$};
         \node[latent,above=of mpparam,yshift = -0.25cm,xshift=0.75cm] (option) {$\option$}; %
         \node[obs,above=of mpparam,xshift=-0.75cm,fill] (context) {$\context$}; %
          
         \edge {context} {option}
         \edge {option,context} {mpparam}  }}
         \subcaption[]{}
         \label{fig::pgm_inference}
   \end{minipage}\hfill
   \begin{minipage}[b]{0.2\textwidth}
        \centering
        \resizebox{0.7\textwidth}{!}{
        \tikz{
         \node[latent] (mpparam) {$\mpparam$};
         \node[latent,above=of mpparam,yshift = -0.25cm,xshift=0.75cm] (context) {$\context$}; %
         \node[obs,above=of mpparam,xshift=-0.75cm,fill] (option) {$\option$}; %
         
         \edge {option} {context}
         \edge {option,context} {mpparam}  }}
         \subcaption[]{}
         \label{fig::pgm_training}
   \end{minipage}\hfill
   \begin{minipage}[b]{0.3\textwidth}
        \centering
        \resizebox{0.93\textwidth}{!}{\includegraphics[trim=0in 0.4in 0in 0.4in, clip]{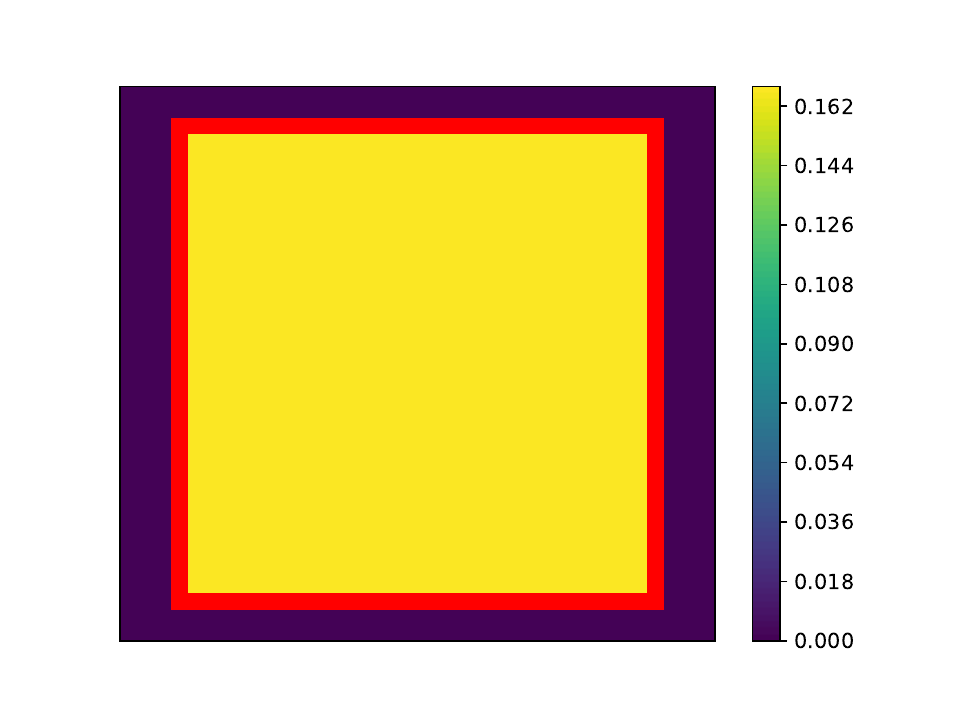}}
        \subcaption[]{}
        \label{fig::vsl::per_expert_context_distr_illustration}
   \end{minipage}\hfill
   \begin{minipage}[b]{0.3\textwidth}
        \centering
        \resizebox{0.93\textwidth}{!}{\includegraphics[trim=0in 0.4in 0in 0.4in, clip]{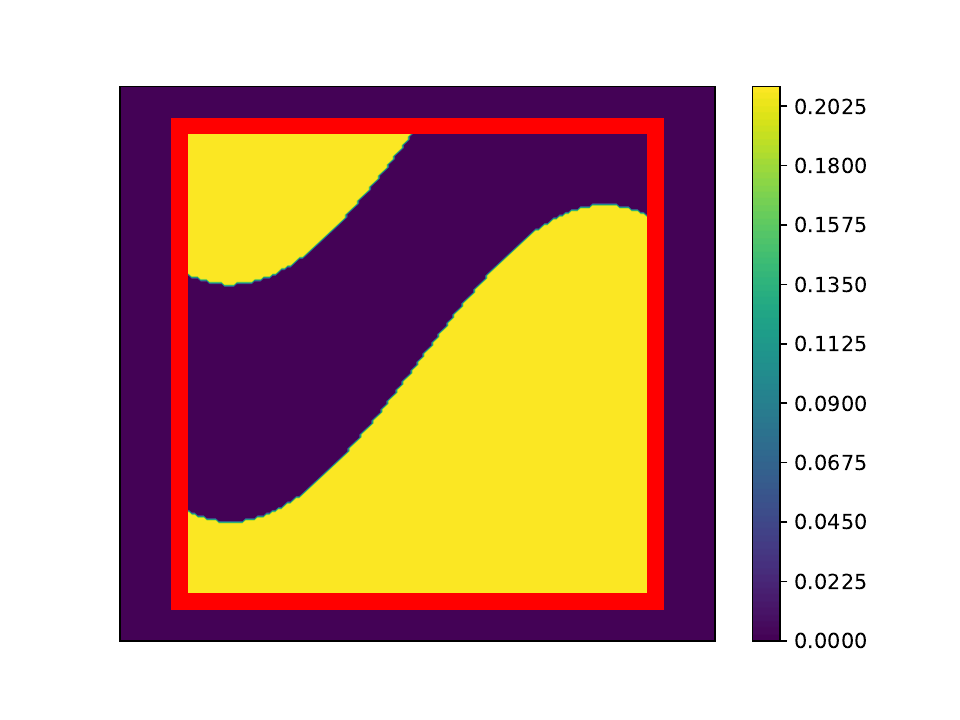}}
        \subcaption[]{}
        \label{fig::vsl::env_context_distr_illustration}
   \end{minipage}\hfill
   \caption{Probabilistic Graphical Models (PGMs) during \textbf{inference} \textbf{a)} and \textbf{training} \textbf{b)}. During \textbf{a))} the model observes the contexts $\context$ from the environment. An expert $\option$ is sampled from $\gating{\pi}$, which leads to an adjustment of the motion primitive parameters $\mpparam$ by $\expert{\pi}$. We iterate over each expert during \textbf{(b)}, sample the contexts $\context$ and $\mpparam$ from the per-expert distribution $\contextcomp{\pi}$ and $\expert{\pi}$ respectively. Sampling from $\contextcomp{\pi}$ allows shaping the expert's curriculum. \textbf{c)} illustrates the environment's context distribution $\contextdistr{p}$ and a possibly optimal $\contextcomp{\pi}$ (\textbf{d))} in two-dim. space. Yellow areas indicate high and purple zero probability. The illustrations show that optimizing $\contextcomp{\pi}$ requires dealing with i) step-like non-linearities, ii) multi-modality, iii) bounded within the red rectangle support of $\contextdistr{p}$, complicating exploration. \vspace{-0.5cm}}
\end{figure*}

\section{Additional Information to Self-Paced Diverse Skill Learning with MoE}\label{sec::Appendix::Self-Paced_Sill_Learning}

The general self-paced diverse skill learning objective

\begin{align*}
    \max_{\policy{\pi},\contextdistr{\pi}} & \mathbb{E}_{\contextdistr{\pi}}\left[\mathbb{E}_{\policy{\pi}} \left[\reward\right] + \alpha \textrm{H}\left[\policy{\pi}\right]\right] -\beta\KL{\contextdistr{\pi}}{\contextdistr{p}}
\end{align*}

can be reformulated to 
\begin{align}
    \label{eq::maxEntrObj}
    \max_{\joint{\pi}} &\mathbb{E}_{\mgating{\pi}, \contextcomp{\pi}}\left[\mathbb{E}_{\expert{\pi}} \left[\reward + \alpha \log{\responsibility{\pi}}\right] + \beta \log{\contextdistr{p}} + (\beta- \alpha) \log \gating{\pi}\right]  \nonumber \\ & + \alpha \mathbb{E}_{\mgating{\pi}, \contextcomp{\pi}}\left[\textrm{H}\left[\expert{\pi}\right]\right] + \beta \mathbb{E}_{\mgating{\pi}}\left[\textrm{H}\left[\contextcomp{\pi}\right]\right] + \beta \textrm{H}\left[\mgating{\pi}\right],
\end{align}
by inserting $\policy{\pi}$, $\contextdistr{\pi}$ from Eq. (\ref{eq::MixtureModel}) into Eq. (\ref{eq::maxEntrObj}) and applying Bayes theorem. This objective is not straightforward to optimize for Mixture of Experts MoE models and requires further steps to introduce a lower bound  (see Section \ref{sec::Preliminaries}) that can be efficiently optimized. Please note that the variational distributions in Eq. \ref{eq::LBComps_with_aux} and Eq. \ref{eq::LBCtxtDistr_with_aux} can be calculated in closed form by the identities
\begin{align*}
   \tilderesponsibility{\pi} &=  \pi_{old}(o|\context, \mpparam) = \frac{\pi_{old}(\mpparam|\context, o)\pi_{old}(o|\context)}{\pi_{old}(\mpparam|\context)} \\
   \tildegating{\pi} &= \pi_{old}(o|\context) = \frac{\pi_{old}(\context|o) \pi(o)}{\pi_{old}(\context)} 
\end{align*}
We refer the interested reader to \cite{celik2022specializing} for a detailed derivation.

\section{Additional Related Work} \label{sec::Appendix::RelatedWork}
\textbf{Unsupervised Reinforcement Learning.} Another field of research that considers learning diverse policies is unsupervised reinforcement learning (URL). In URL the agent is first trained solely with an intrinsic reward to acquire a diverse set of skills from which the most appropriate is picked to solve a downstream task. More related to our work is a group of algorithms that obtain their intrinsic reward based on information-theoretic formulations \citep{urlb, eysenbach2018diversity, CamposTXSGT20, lee2019efficient, aps-liu21b}. However, their resulting objective is based on the mutual-information and differs from the objective we maximize. The learned skills in the pre-training aim to cover distinct parts of the state-space during pre-training in the absence of an extrinsic task reward which implies that skills are not explicitly trained to solve the same task in different ways. Those methods operate within the step-based RL setting which differs from CEPS.

\section{Additional Information to Diverse Skill Learning}\label{sec::Appendix::additional_information}

\subsection{The Parameterization of the Mixture of Experts (MoE) Model}
In the following, we provide details on the parameterization of the MoE model.

\textbf{Parametrization of the expert $\expert{\pi}$.}
We parameterize each expert $\expert{\pi}$ as a Gaussian policy $\mathcal{N}(\cvec{\mu}_{\cvec \gamma}(\context), \cvec{\Sigma}_{\cvec \gamma}(\context)$, where the mean $\cvec{\mu}_{\cvec \gamma}(\context)$ and the covariance $\cvec{\Sigma}_{\cvec \gamma}(\context)$ are functions of the context $\context$ and parameterized by a neural network with parameters $\cvec \gamma$. 
Although the covariance $\cvec{\Sigma}_{\cvec \gamma}(\context)$ is formalized as a function of the context $\context$, we have not observed any advantages in doing so. In our experiments, we therefore parameterize the covariance as a lower-triangular matrix $\cvec{L}$ and form the covariance matrix $\cvec{\Sigma} = \cvec{LL}^T$.

\textbf{Parameterization of the per-expert context distribution $\contextcomp{\pi}$}. The reader is referred to Section \ref{sec::VSL} for details on the parameterization of $\contextcomp{\pi}$

\textbf{Parameterization of the prior $\mgating{\pi}$}. We fix the prior $\mgating{\pi}$ to a uniform distribution over the number \textit{K} of available components and do not further optimize this distribution. This is a useful definition to increase the entropy of the mixture model. 

\textbf{Parameterization of the context distribution $\contextdistr{\pi}$.} Due to the relation $\contextdistr{\pi} = \sum_o \contextcomp{\pi} \mgating{\pi}$, $\contextdistr{\pi}$ is defined by $\contextcomp{\pi}$ and does not need explicit modelling.

\textbf{Parameterization of the gating distribution $\gating{\pi}$}. Due to the relation $\gating{\pi}=\frac{\contextcomp{\pi}\mgating{\pi}}{\contextdistr{\pi}}$ we do not need an explicit parameterization of $\gating{\pi}$ and can easily calculate the probabilities for choosing the expert $o$ given a context $\context$.

\subsection{Using Motion Primitives in the Context of Reinforcement Learning}
Motion Primitives (MPs) are a low-dimensional representation of a trajectory. For instance, instead of parameterizing a desired joint-level trajectory as the single state in each time step, MPs introduce a low-dimensional parameter vector $\mpparam$ which concisely defines the trajectory to follow. The generation of the trajectory depends on the method that is used. Probabilistic Movement Primitives (ProMPs) \cite{paraschos2013probabilistic} for example define the desired trajectory as a simple linear function $\cvec \tau = \cvec \Phi ^T \mpparam$, where $\cvec \Phi$ are time-dependent basis functions (e.g. normalized radial basis functions). Dynamic Movement Primitives (DMPs) \cite{schaal2006dynamic} rely on a second-order dynamic system that provides smooth trajectories in the position and velocity space. Recently Probabilistic Dynamic Movement Primitives (ProDMPs) were introduced by \citet{li2023prodmp} and combines the advantages of both methods, that is the easy generation of trajectories and smooth trajectories. We therefore rely on ProDMPs throughout this work.

In the context of reinforcement learning, the policy $\policy{\pi}$, or in our case an expert $\expert{\pi}$ defines a distribution over the parameters $\mpparam$ of the MP depending on the observed context $\context$. This allows the policy to quickly adapt to new tasks defined by $\context$.

\subsection{Algorithm Details}
Detailed descriptions of the algorithm during training and during inference are provided in the algorithm boxes Alg. \ref{algo_training} and Alg. \ref{algo_training}, respectively. In each iteration during training, we sample a batch of contexts $\context$ from the environment by resetting it. We then iterate over each expert and evaluate the probabilities of these contexts $\context$ on each per-expert context distribution $\contextcomp{\pi}$ and sample then training contexts $\context_T$ from them. From the corresponding expert $\expert{\pi}$ we sample motion primitive parameters $\mpparam$ and evaluate the samples ($\context_T, \mpparam$) on the environment and observe a return $\reward$ which we use to update the experts $\expert{\pi}$ and the per-expert context distributions $\contextcomp{\pi}$ by maximizing Obj. \ref{eq::LBComps_with_aux_with_TRConstraint} and Obj. \ref{eq::LBCtxtDistr_with_aux_EBM} respectively. During inference, we observe contexts $\context$ from the environment, calculate the gating distributions $\gating{\pi} = \frac{\contextcomp{\pi} \mgating{\pi}}{\contextdistr\pi}$ from which we sample the expert $o$. We then either take the mean or sample an $\mpparam$ from this expert and execute it on the environment.
\begin{algorithm}[H] 
	\caption{Di-SkilL Training}
	\begin{algorithmic}[1]
		\renewcommand{\algorithmicrequire}{\textbf{Input:}}
		\renewcommand{\algorithmicensure}{\textbf{Output:}}
		\REQUIRE $\alpha, ~ \beta$, N(max. iterations), K(num. experts),T(num. samples per expert)
		\ENSURE  $\policy{\pi}$
		\FOR{$k = 1$ to N}
		\STATE $\context\sim\contextdistr{p}$ (context batch by environment resetting)
		\FOR{$ o = 1$ to $K$}
            \STATE $\context_{T} \sim \contextcomp{\pi}$ (context batch from EBM)
            \STATE $\mpparam \sim \pi(\mpparam|\context_T, o)$
            \STATE $\reward \leftarrow$ eval($\mpparam, \context_T$)
            \STATE $\expert{\pi} \leftarrow$ Obj. \ref{eq::LBComps_with_aux_with_TRConstraint}
            \STATE $\contextcomp{\pi} \leftarrow$ Obj. \ref{eq::LBCtxtDistr_with_aux_EBM}
		\ENDFOR
		\ENDFOR
	\end{algorithmic}
	\label{algo_training}
\end{algorithm}
\begin{algorithm}[H]
	\caption{Di-SkilL Inference}
	\begin{algorithmic}[1]
		\renewcommand{\algorithmicrequire}{\textbf{Input:}}
		\renewcommand{\algorithmicensure}{\textbf{Output:}}
		\REQUIRE $\policy{\pi}$
        \STATE $\context\sim\contextdistr{p}$ (observe contexts from environment)
        \STATE $o \sim \gating{\pi}$, where $\gating{\pi}=\frac{\contextcomp{\pi}\mgating{\pi}}{\contextdistr{\pi}}$
        \STATE $\mpparam \sim \pi(\mpparam|\context, o)$
        \STATE $\reward \leftarrow$ eval($\mpparam, \context$)
	\end{algorithmic}
	\label{algo_inference}
\end{algorithm}

\section{Experimental Details}\label{sec::Appendix:ExpDetails}
\subsection{Environment Details}

\subsubsection{Table Tennis Easy}\label{sec::Appendix:AblationStudies}
\paragraph{Environment.} We use the same table tennis environment as presented in \citep{otto2023deep}, in which a 7 Degree of Freedom (DoF) robot has to return a ball to a desired ball landing position. The \textbf{context} is the four-dimensional space of the ball's initial landing position ( $x \in [-1, -0.2]$, $y \in [-0.65, 0.65]$) on the robot's table side and the desired ball landing position ($x \in [-1.0, -0.2]$, $y \in [-0.6, 0.6]$) on the opponent's table side. The robot is controlled with torques on the joint level in each time step. The torques are generated by the tracking controller (PD-controller) that tracks the desired trajectory generated by the motion primitive. We consider three basis functions per joint resulting in a 21-dimensional parameter ($\mpparam$) space. We additionally allow the agent to learn the trajectory length and the starting time step of the trajectory. Note that the starting point allows the agent to define when after the episode's start the generated desired trajectory should be tracked. Induced by the varying contexts, this is helpful to react to the varying time the served ball needs to reach a positional space that is convenient to hit the ball with the robot's racket. Overall the \textbf{parameter space} is 23 dimensional. The task is considered successful if the returned ball lands on the opponent's side of the table and within $\leq 0.2$m to the goal location.

The \textbf{reward function} is unchanged from \citep{otto2023deep} and is defined as
\begin{equation*}
    R_{task} = \begin{cases}
        0, & \text{if cond. 1}, \\
        f_2(\mathbf{p}_{r}, \mathbf{p}_{b}) & \text{if cond. 2}, \\
        f_3(\mathbf{p}_{r}, \mathbf{p}_{b}, \mathbf{p}_l, \mathbf{p}_{goal}) &\text{if cond. 3},\\
        f_4(\mathbf{p}_{r}, \mathbf{p}_{b}, \mathbf{p}_l, \mathbf{p}_{goal}) &\text{if cond. 4},\\
        f_5(\mathbf{p}_{r}, \mathbf{p}_{b}, \mathbf{p}_l, \mathbf{p}_{goal}) &\text{if cond. 5},\\
        \end{cases}
\end{equation*}
where $\mathbf{p}_r$ is the executed trajectory position of the racket center, $\mathbf{p}_b$ is the executed position trajectory of the ball, $\mathbf{p}_{l}$ is the ball landing position, $\mathbf{p}_{goal}$ is the target position. The individual functions are defined as 
\begin{align*}
    f_2(\mathbf{p}_{r}, \mathbf{p}_{b}) &= 0.2-0.2g(\mathbf{p}_{r}, \mathbf{p}_{b})\\
    f_3(\mathbf{p}_{r}, \mathbf{p}_{b}, \mathbf{p}_l, \mathbf{p}_{goal}) &= 3-2g(\mathbf{p}_{r}, \mathbf{p}_{b})-h(\mathbf{p}_{l}, \mathbf{p}_{goal})\\
    f_4(\mathbf{p}_{r}, \mathbf{p}_{b}, \mathbf{p}_l, \mathbf{p}_{goal}) &= 6-2g(\mathbf{p}_{r}, \mathbf{p}_{b})-4h(\mathbf{p}_{l}, \mathbf{p}_{goal})\\
    f_5(\mathbf{p}_{r}, \mathbf{p}_{b}, \mathbf{p}_l, \mathbf{p}_{goal}) &= 7-2g(\mathbf{p}_{r}, \mathbf{p}_{b})-4h(\mathbf{p}_{l}, \mathbf{p}_{goal}),
\end{align*}
where $g(\cvec x,\cvec y) = \tanh{(\min||\cvec x-\cvec y||^2)}$ and $h(\cvec x,\cvec y) = \tanh{(||\cvec x-\cvec y||^2)}$.
The different conditions are
\begin{itemize}
    \item cond. 1: the end of the episode is not reached,
    \item cond. 2: the end of the episode is reached,
    \item cond. 3: cond.2 is satisfied and the robot did hit the ball,
    \item cond. 4: cond.3 is satisfied and the returned ball lands on the table,
    \item cond. 5: cond.4 is satisfied and the landing position is at the opponent's side.
\end{itemize}
The episode ends when any of the following conditions are met
\begin{itemize}
    \item the maximum horizon length is reached
    \item ball did land on the floor without hitting
    \item ball did land on the floor or table after hitting
\end{itemize}

The whole desired trajectory is obtained ahead of environment interaction, making use of this property we can collect some samples without physical simulation. The reward function based on this desired trajectory is defined as
\begin{equation*}
    r_{traj} = -\sum_{(i,j)}{|\tau_{ij}^d| - |q^b_j|} , \quad 
    (i,j) \in \{(i,j)\mid |\tau_{ij}^d| > |q^b_j|\}
\end{equation*}
where $\tau^d$ is the desired trajectory, $i$ is the time index, $j$ is the joint index, $q^b$ is the joint position upper bound. The desired trajectory is considered as invalid if $r_{traj} < 0$, an invalid trajectory will not be executed on the robot. The overall reward is defined as:
\begin{equation*}
    r = \begin{cases}
    r_{traj}, & r_{traj} < 0 \\
    r_{task}, & \text{otherwise}
    \end{cases}
\end{equation*}
 
\paragraph{SVSL.} SVSL requires designing a guiding punishment term for context samples that are not in a valid region. For the four-dimensional context space in table tennis, this can be done using quadratic functions (as proposed in the original work \citep{celik2022specializing}): 
\begin{align*}
    R_c(\context) = -20 \cdot d_c^2,
\end{align*}
where $d_c^2$ is the distance of the current context $\cvec c$ to the valid context region.

\paragraph{SVSL Hyperparameters} All hyperparameters are summarized in the Table \ref{tab:hypparams_svsl}.

\textbf{Hyperparameters} are listed in the Table \ref{tab:hyperparams_ablation_bbrl_ours}.

\subsubsection{Table Tennis Task Hard}

\paragraph{Environment.} We extend the table tennis environment described in Appendix \ref{sec::Appendix:AblationStudies} by additionally including the ball's initial velocity in the context space making the task harder as the agent has to react to ranging velocities now. We define the initial velocity $v_x \in [1.5\frac{m}{s}, 4 \frac{m}{s}]$. Note that every single constellation within the resulting context space is a valid context. However, there exist ball landing positions that can not be set along with a subset of the initial velocity range. This makes designing a guiding punishment term for SVSL especially difficult. We adopt the \textbf{parameter space} and the \textbf{reward function} as defined in the standard table tennis environment as described in Appendix \ref{sec::Appendix:AblationStudies}.

\textbf{Hyperparameters} are listed in the Table \ref{tab:hyperparams_extended_TT_bbrl_ours}.

\subsection{Hopper Jump}
\textbf{Environment.} We use the same hopper jump environment as presented in \cite{otto2023deep}, in which the hopper \cite{gym} has to jump as high as possible and land at a specified position. The \textbf{context} is the four-dimensional space of the last three joints of the hopper and the goal landing position $[j_3, j_4, j_5, g]$, where the ranges are from $[-0.5, -0.2, 0, 0.3]$ to $[0.0, 0.0, 0.785,1.35]$. The hopper is controlled the same as in \cite{gym}. Here, we consider three basis functions per joint and a goal basis resulting in a \textbf{parameter space} ($\mpparam$) of 12 dimensions. The reward is non-markovian and is unchanged from \cite{otto2023deep}. 

In each time-step $t$ the action cost 
\begin{align*}
    \tau_t = 10^{-3}\sum_i^K (a^i_t)^2,
\end{align*}
is provided. The variable $K=3$  corresponds to the number of degrees of freedom. At the end of the episode, a reward containing retrospective information about the maximum height in the z-direction of the center of mass achieved $h_{\text{max}}$, the goal landing position of the heel $p_{\text{goal}}$, the foot's heel position when having contact with the ground after jumping the first time $p_{\text{foot, contact}}$ is given. Additionally, per-time information such as $p_{\text{foot, t}}$ describing the position of the foot's heel in world-coordinates is given. The resulting reward function is 
\begin{align*}
    R_{tot} = -\sum_{t=0}^T\tau_t + R_{height} + R_{gdist} + R_{cdist} + R_{healthy},
\end{align*}
where
\begin{align*}
    R_{height} &= 10h_{max}, \\
    R_{gdist} &= ||p_{foot, T} - p_{goal}||_2,\\
    R_{cdist} &= ||p_{foot,contact} - p_{goal}||_2,\\
    R_{healthy} &= \left\{\begin{array}{ll} 2 & \textrm{if } z_T \in [0.5, \infty] \textrm{and } \theta, \gamma, \phi \in [-\infty, \infty]\\
                                           0 & \textrm{else.} \end{array}\right.
\end{align*}
The healthy reward is the same as provided by \cite{gym}.

\textbf{Hyperparameters} are listed in the Table \ref{tab:hyperparams_hopper_jump}.
\subsubsection{Box Pushing with Obstacle Task}

\paragraph{Environment.} We increase the difficulty of the box pushing environment as presented in \citep{otto2023deep}, by changing major parts of the context space. The goal of the box pushing task is to move a box to a specified goal location and orientation using the seven DoF Franka Emika Panda. The newly \textbf{context space} (compared to the original version in \citep{otto2023deep}) are described in the following. We increase the box' goal position range to $x_g \in [0.3, 0.6], y_g \in [-0.7, 0.45]$, and keep the goal orientation angle $\phi \in [0rad, 2\pi rad]$. Additionally, we include an obstacle between the initial box and the box's goal. The range of the obstacle position is $x_o \in [0.3, 0.6], y_o \in [-0.3, 0.15]$. Note that we guarantee a distance of at least 0.15m between the obstacle's position and the initial position as well as at least 0.15m between the obstacle's position and the box's goal position. 

The robot is controlled via torques on the joint level. We use four basis functions per DoF, resulting in a \textbf{parameter space} of 28 dimensions. We consider an episode successful if the box's orientation around the z-axis error is smaller than 0.5 rad and the position error is smaller than 0.05m. 

The \textbf{sparse-in-time reward function} is up to a scaling parameter the same as presented in \citep{otto2023deep}. We describe the whole reward function in the following. 

The box's distance to the goal position is 
\[
 R_\text{goal} = \lVert\mathbf{p}-\mathbf{p}_{goal}\rVert,
\]
where $\mathbf{p}$ is the box position and $\mathbf{p}_{goal}$ is the goal position. The rotation distance is defined as 
\[
R_\text{rotation} =\frac{1}{\pi} \arccos{|\mathbf{r}\cdot\mathbf{r}_{goal}|},
\]
where $\mathbf{r}$ and $\mathbf{r}_{goal}$ are the box orientation and goal orientation quaternion respectively. The incentive to keep the rod within the box is defined as 
\[
R_\text{rod} = \text{clip}(||\mathbf{p}-\mathbf{h}_{pos}||, 0.05, 10),
\]
where $\mathbf{h}_{pos}$ is the position of the rod tip. Similarly, to incentivize to maintain the rod in a desired rotation, the reward
\[
R_\text{rod\_rotation} = \text{clip} (\frac{2}{\pi} \arccos{|\mathbf{h}_{rot}\cdot\mathbf{h}_0|} , 0.25, 2)
\]
is defined, where $\mathbf{h}_{rot}$ and $\mathbf{h}_{0} = (0, 1, 0, 0)$ are the current and desired rod orientation in quaternion respectively. To incentivize the robot to stay within the joint and velocity bounds, the error
\begin{align*}
    \text{err}(\mathbf{q}, \mathbf{\dot{q}}) &= \sum_{i \in \{i | |q_i| > |q_i^{b}|\}}{(|q_i|-|q^{b}_i|)} \\
    &+ \sum_{j \in \{j | |\dot{q}_j| >|\dot{q}_j^{b}|\}}{(|\dot{q}_j|-|\dot{q}^{b}_j|)}
\end{align*}

is used, where $\mathbf{q}$, $\mathbf{\dot{q}}$, $\mathbf{q}^{b}$, and $\mathbf{\dot{q}}^{b}$ are the robot's joint positions and velocities as well as their respective bounds.
To learn low-energy motions, the per-time action (torque) cost
\begin{align*}
    \tau_t = \sum_i^K (a^i_t)^2,
\end{align*}
is used. The resulting temporal sparse reward is given as 
\begin{equation*}
    R_\text{tot} = \begin{cases}
        - R_\text{rod}
        - R_\text{rod\_rotation}
        - 0.02\tau_t 
        - \text{err}(\mathbf{q}, \mathbf{\dot{q}})
        &t<T, 
        \\
        - R_\text{rod}
        - R_\text{rod\_rotation}
        - 0.02\tau_t 
        - \text{err}(\mathbf{q}, \mathbf{\dot{q}})\\
        -350 R_\text{goal}
        -200R_\text{rotation}
        &t=T,
        \end{cases}
\end{equation*}
where $T=100$ is the horizon of the episode. The reward gives relevant information to solve the ask only in the last time step of the episode, which makes exploration hard. 

\paragraph{Further Visualizations of learned skills.} We show additional plots of the box's trajectories in the box pushing task in Fig. \ref{fig::exps::diverse_box_trajectories_appendix}.

\textbf{Hyperparameters} are listed in the Table \ref{tab:BoxP_bbrl_ours}.

\subsection{Extended 5-Link Reacher Task}
\paragraph{Environment.} In the 5-Link Reacher task, a 5-link planar robot has to reach a goal position with its tip. The reacher's initial position is straight to the right. This task is difficult to solve, as it introduces multi-modality in the behavior space.  \citep{otto2023deep} avoided this multi-modality by constraining the y coordinate of the goal position to $y\geq0$, i.e. the first two quadrants. We adopt the 5Link-Reacher task by increasing the context space to the full space, i.e. all four quadrants. We consider 5 basis functions per joint leading to a 25-dimensional \textbf{parameter space}. We consider the \textbf{sparse reward function} presented in \citep{otto2023deep} as
\begin{equation*}
    R_\text{tot} = \begin{cases}
        - \tau_t 
        &t<T, 
        \\
        - \tau_t 
        -200 R_\text{goal}
        -10 R_\text{vel}
        &t=T,
        \end{cases}
\end{equation*}
where 
\[
 R_\text{goal} = \lVert \mathbf{p}-\mathbf{p}_{goal}\rVert_2
\]
and 
\begin{align*}
    \tau_t = \sum_i^K (a^i_t)^2.
\end{align*}
The sparse reward only returns the task reward in the last time step T and additionally adds a velocity penalty $R_\text{vel} = \sum_i^K (\dot{q}_T^i)^2$. The joint velocities are denoted as$\mathbf{\dot{q}}$. This velocity penalty avoids overshooting in the last time step.

\textbf{Hyperparameters} are listed in the Table \ref{tab:Reacher_bbrl_ours}.

\subsection{Robot Mini Golf Task}
\paragraph{Environment.} In the robot mini golf task the agent needs to hit a ball while avoiding the two obstacles, such that it passes the tight goal to achieve a bonus. The \textbf{context space} consists of the ball's initial x-position $x_{ball}\in[0.25m, 0.6m]$, the XY positions of the green obstacle $x_{obs}\in [0.3, 0.6]$ and $y_{obs}\in[-0.5, -0.1]$ and the x positions of the goal $x_{ball}\in[0.25, 0.6]$. The \textbf{parameter space} is 29 dimensional resulting from the 4 basis functions per joint and an additional duration parameter which allows the robot to learn the duration of the trajectory. The robot starts always at the same position. The \textbf{reward} function consists of three stages: 
\begin{equation*}
    R_{task} = \begin{cases}
        -0.0005 \cdot \tau_t & \text{if cond. 1}, \\
        0.2-0.2\tanh{(\min{||\mathbf{p}_{r}-\mathbf{p}_{b}||})} & \text{if cond. 2}, \\
        2-2\tanh{(\min{||\mathbf{p}_{b}-\mathbf{p}_{g}||})}\\
        -\tanh{(||\mathbf{p}_{b,y}-\mathbf{p}_{thresh,y}||)} &\text{if cond. 3},\\
        6 &\text{if cond. 3} ,
        \end{cases}
\end{equation*}
where the individual conditions are
\begin{itemize}
    \item cond. 1: the end of the episode is not reached,
    \item cond. 2: the end of the episode is reached and the robot did not hit the ball,
    \item cond. 3: the end of the episode is reached and the robot has hit the ball, but the ball didn't pass the goal
    \item cond. 4: the end of the episode is reached, robot has hit the ball and the ball has passed the goal for at least 0.75m 
\end{itemize}
The episode ends when the maximum horizon length $T=100$ is achieved. 
We again make use of the advantage that we obtain the whole desired trajectory ahead of the environment interaction, such that we can collect some samples without physical simulation. The reward function based on this desired trajectory is defined as
\begin{equation*}
    r_{traj} = \sum_{(i,j)}{|\tau_{ij}^d| - |q^b_j|} , \quad 
    (i,j) \in \{(i,j)\mid |\tau_{ij}^d| > |q^b_j|\}
\end{equation*}
where $\tau^d$ is the desired trajectory, $i$ is the time index, $j$ is the joint index, $q^b$ is the joint position upper bound. The desired trajectory is considered as invalid if $r_{traj} < 0$, an invalid trajectory will not be executed on the robot. Additionally, we provide a punishment, if the agent samples invalid duration times
\begin{equation*}
    r_{dur} = -3\left(\text{max}(0, t_d-t{d,max}) + \text{max}(0, t_{d,mint} - t_d) \right),
\end{equation*}

where $t_{d,max}=1.7s, t_{d,min} = 0.45s$ and $t_d$ is the duration in seconds chosen by the agent. The overall reward is defined as:
\begin{equation*}
    r = \begin{cases}
    r_{traj}, -20(r_{traj} + r_{dur})-5 \quad  &\text{if invalid duration,} \\
    & \text{or trajectory} \\
    r_{task}, & \text{otherwise.}
    \end{cases}
\end{equation*}
\textbf{Hyperparameters} are listed in the Table  \ref{tab:minigolf_bbrl_ours}.

\section{Additional Evaluations}\label{sec:appendix:additional_evaluations}
\begin{figure*}[t!]
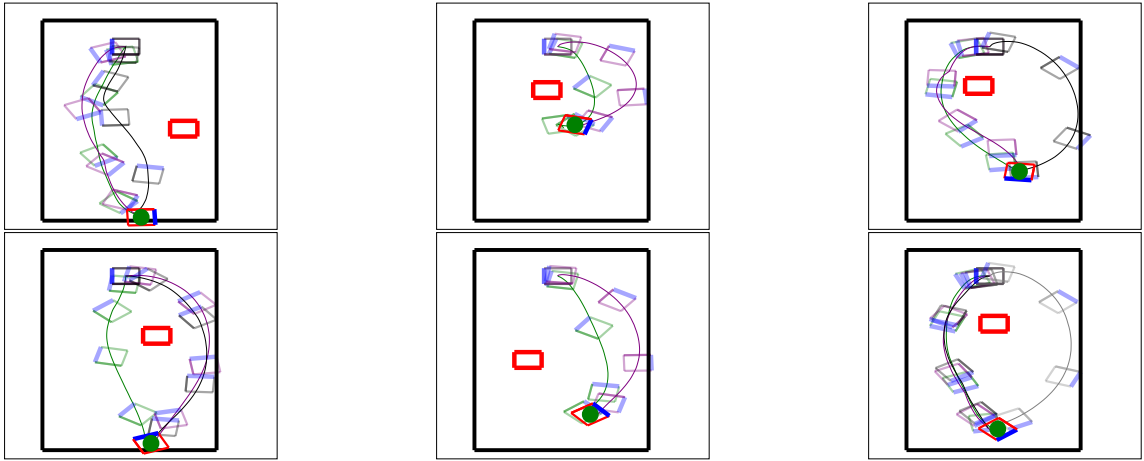

    \begin{minipage}[b]{0.33\textwidth}
        \centering
       \resizebox{0.65\textwidth}{!}{\input{figures/experiments/bp_obs/box_positions_appendix/box_trajs_1}}
       \label{}
   \end{minipage}\hfill
   \begin{minipage}[b]{0.33\textwidth}
        \centering
       \resizebox{0.65\textwidth}{!}{\input{figures/experiments/bp_obs/box_positions_appendix/box_trajs_2}}
       \label{}
   \end{minipage}\hfill
   \begin{minipage}[b]{0.33\textwidth}
        \centering
       \resizebox{0.65\textwidth}{!}{\input{figures/experiments/bp_obs/box_positions_appendix/box_trajs_3}}
       \label{}
   \end{minipage}\hfill
   
   \begin{minipage}[b]{0.33\textwidth}
        \centering
       \resizebox{0.65\textwidth}{!}{\input{figures/experiments/bp_obs/box_positions_appendix/box_trajs_4}}
       \label{}
   \end{minipage}\hfill
   \begin{minipage}[b]{0.33\textwidth}
        \centering
       \resizebox{0.65\textwidth}{!}{\input{figures/experiments/bp_obs/box_positions_appendix/box_trajs_5}}
       \label{}
   \end{minipage}\hfill
   \begin{minipage}[b]{0.33\textwidth}
        \centering
       \resizebox{0.65\textwidth}{!}{\input{figures/experiments/bp_obs/box_positions_appendix/box_trajs_6}}
       \label{}
   \end{minipage}\hfill
   \caption{\textbf{Additional Diverse Skills for the Box Push Obstacle Task learned by Di-SkilL.} We fix the contexts and sample experts which we subsequently execute. This leads to diverse behaviors in the motion primitive parameter space$\mpparam$ which leads to different trajectories of the pushed box on the table.}
   \label{fig::exps::diverse_box_trajectories_appendix}
\end{figure*}

\begin{figure*}[t]
   \begin{minipage}[b]{0.2\textwidth}
        \centering
       \resizebox{\textwidth}{!}{\includegraphics{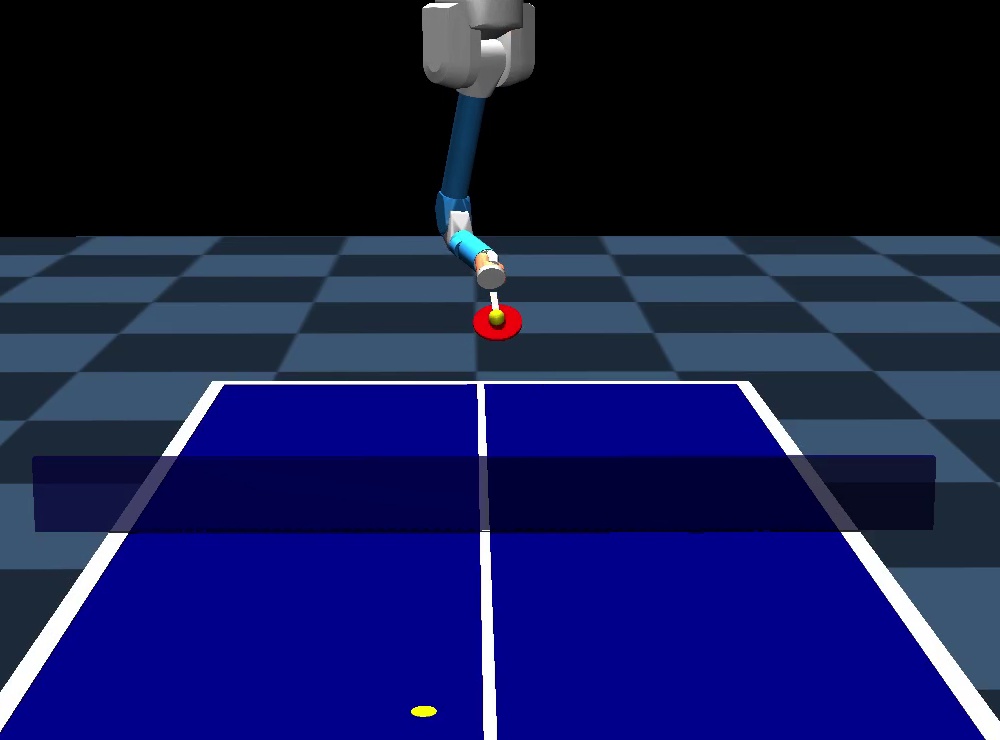}}
       \label{}
    \end{minipage}\hfill
   \begin{minipage}[b]{0.2\textwidth}
        \centering
       \resizebox{\textwidth}{!}{\includegraphics{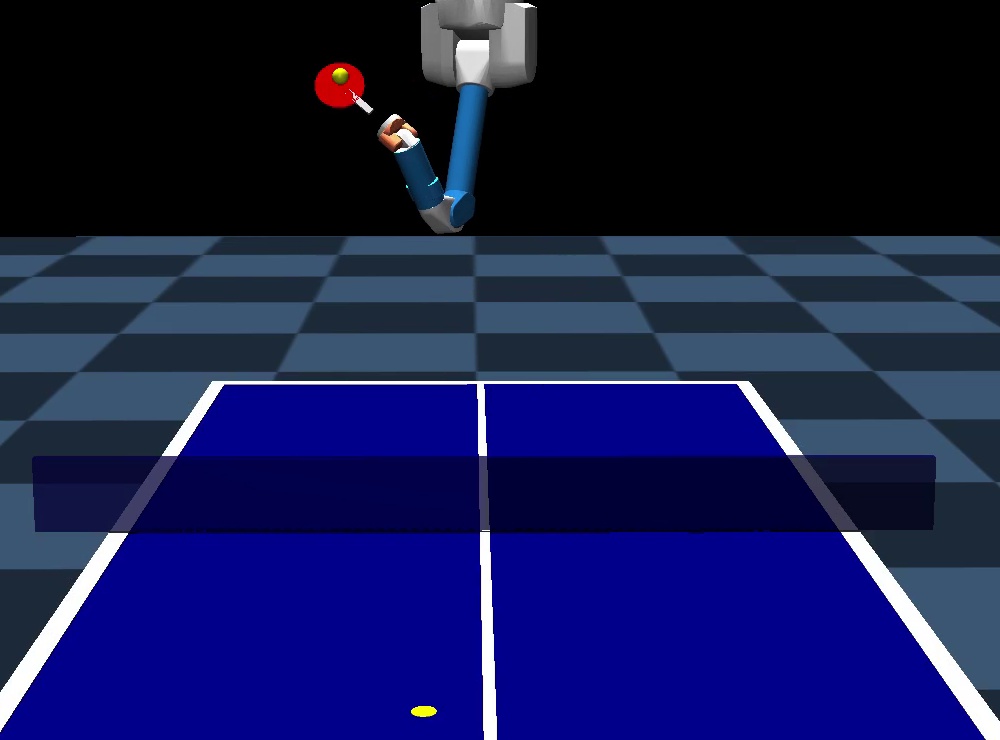}}
       \label{}
    \end{minipage}\hfill
    \begin{minipage}[b]{0.2\textwidth}
        \centering
       \resizebox{\textwidth}{!}{\includegraphics{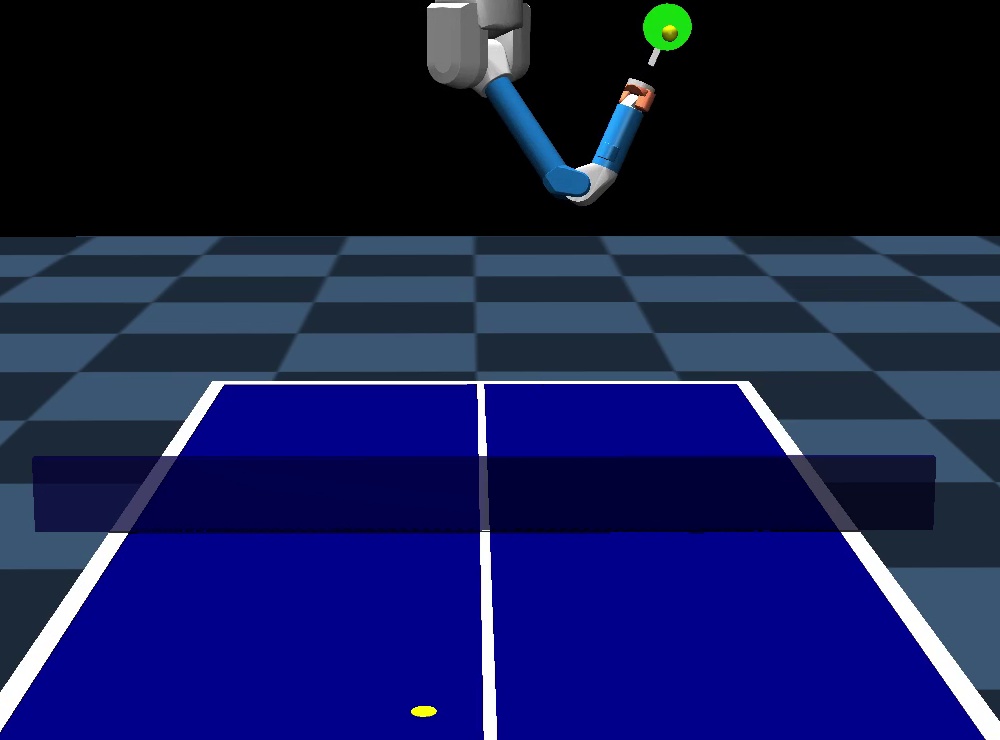}}
       \label{}
   \end{minipage}\hfill
   \begin{minipage}[b]{0.2\textwidth}
        \centering
       \resizebox{\textwidth}{!}{\includegraphics{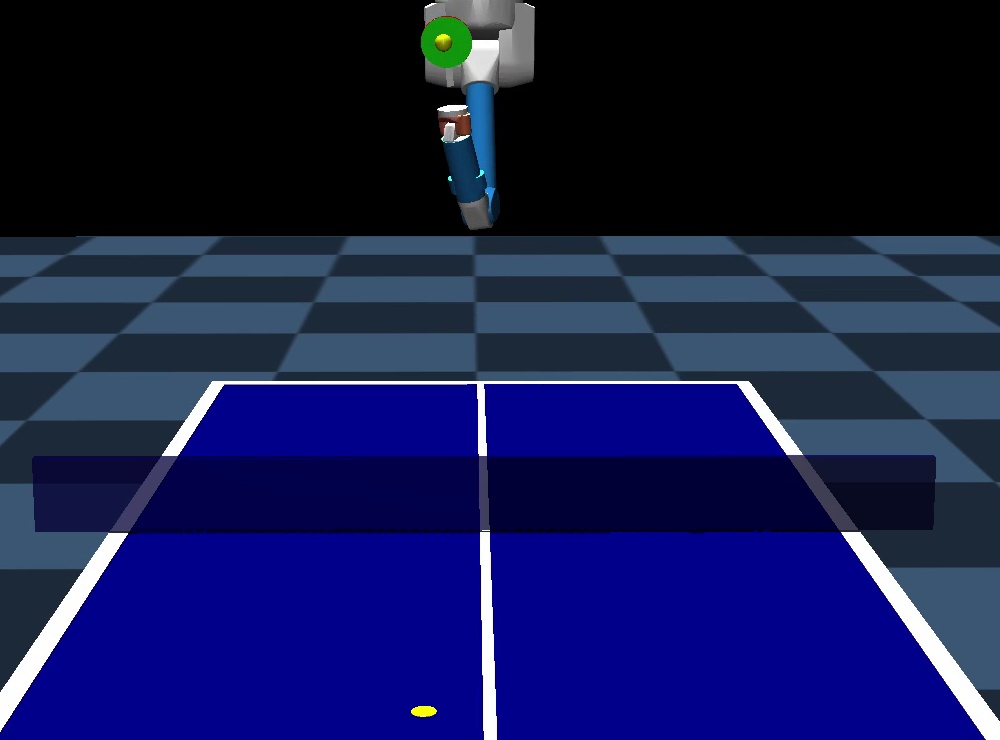}}
       \label{}
   \end{minipage}\hfill
   \begin{minipage}[b]{0.2\textwidth}
        \centering
       \resizebox{\textwidth}{!}{\includegraphics{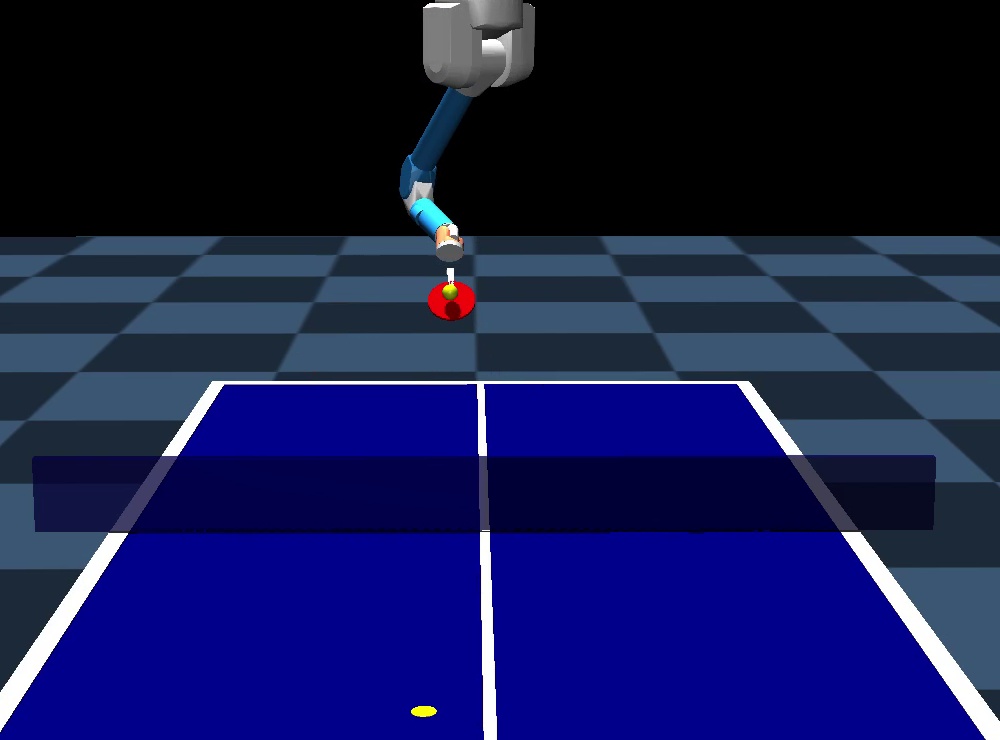}}
       \label{}
   \end{minipage}\hfill

   \begin{minipage}[b]{0.2\textwidth}
        \centering
       \resizebox{\textwidth}{!}{\includegraphics{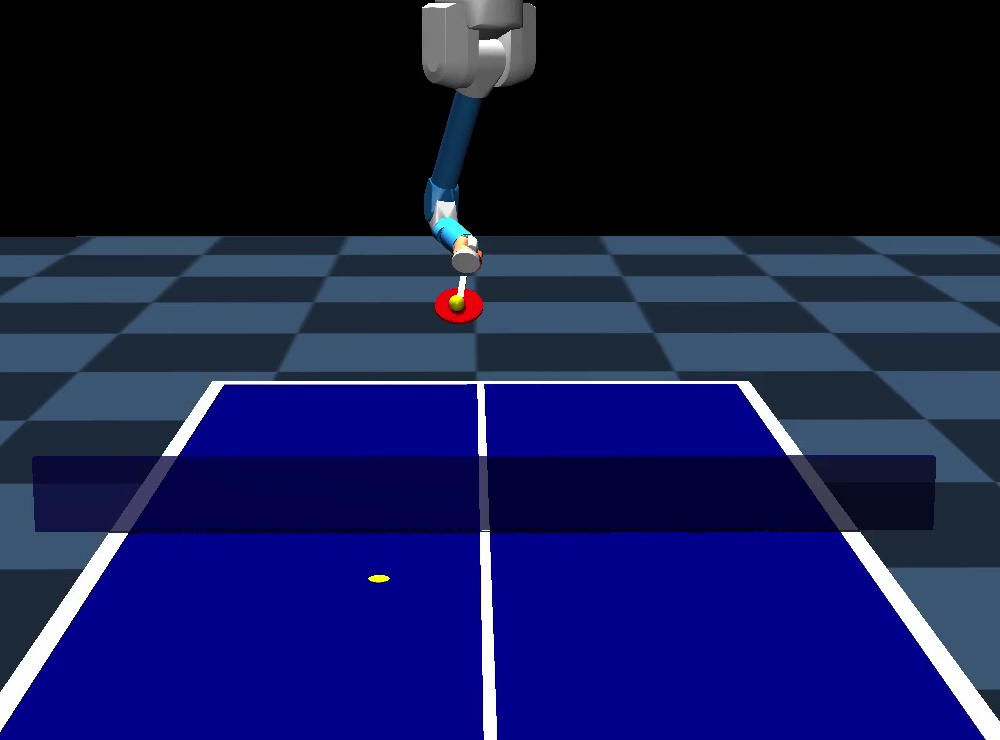}}
       \label{}
    \end{minipage}\hfill
   \begin{minipage}[b]{0.2\textwidth}
        \centering
       \resizebox{\textwidth}{!}{\includegraphics{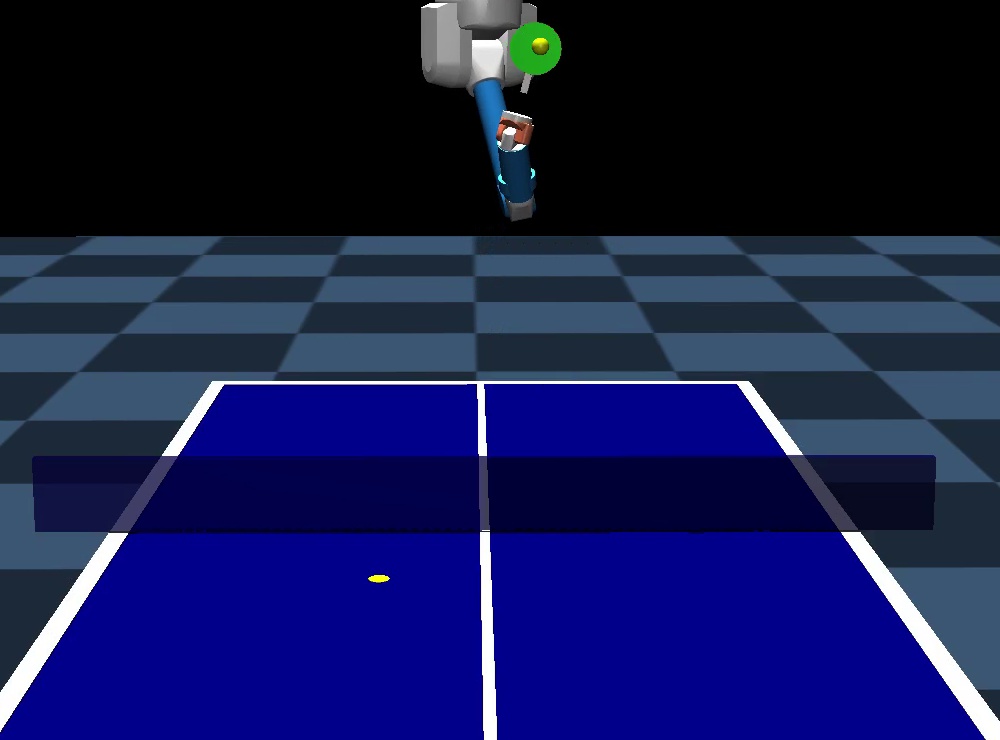}}
       \label{}
    \end{minipage}\hfill
    \begin{minipage}[b]{0.2\textwidth}
        \centering
       \resizebox{\textwidth}{!}{\includegraphics{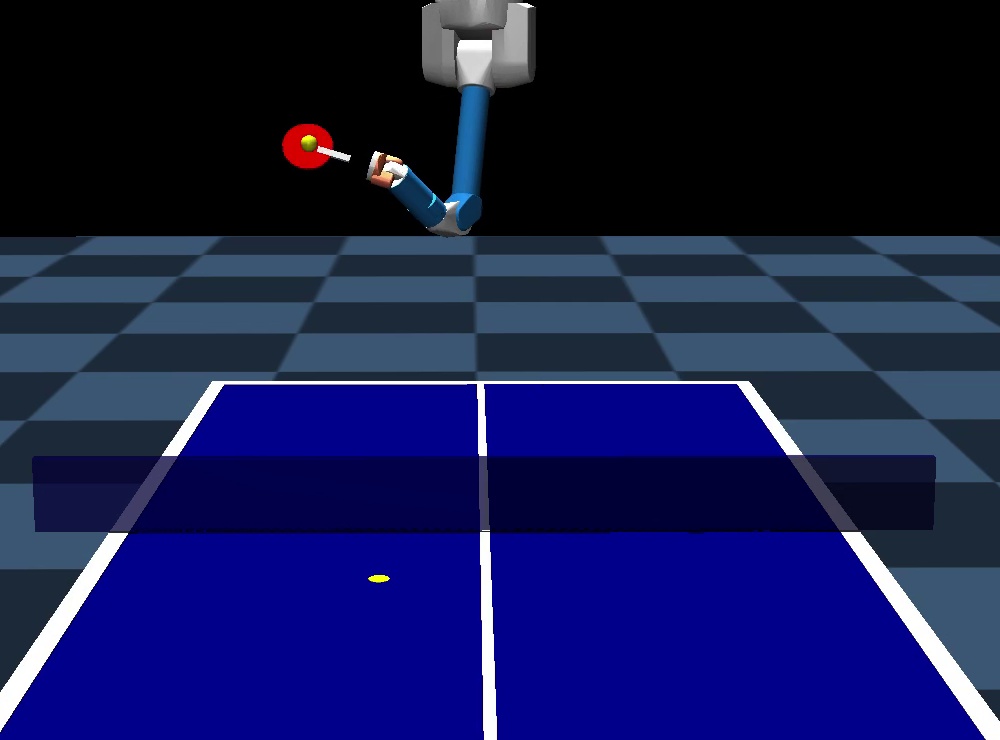}}
       \label{}
   \end{minipage}\hfill
   \begin{minipage}[b]{0.2\textwidth}
        \centering
       \resizebox{\textwidth}{!}{\includegraphics{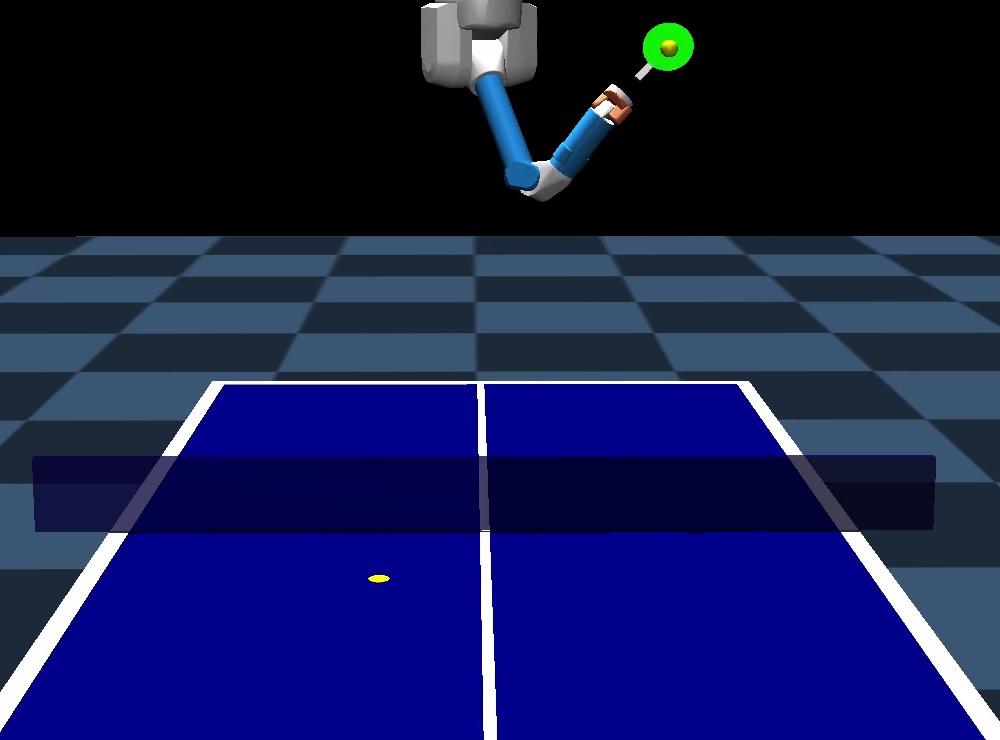}}
       \label{}
   \end{minipage}\hfill
   \begin{minipage}[b]{0.2\textwidth}
        \centering
       \resizebox{\textwidth}{!}{\includegraphics{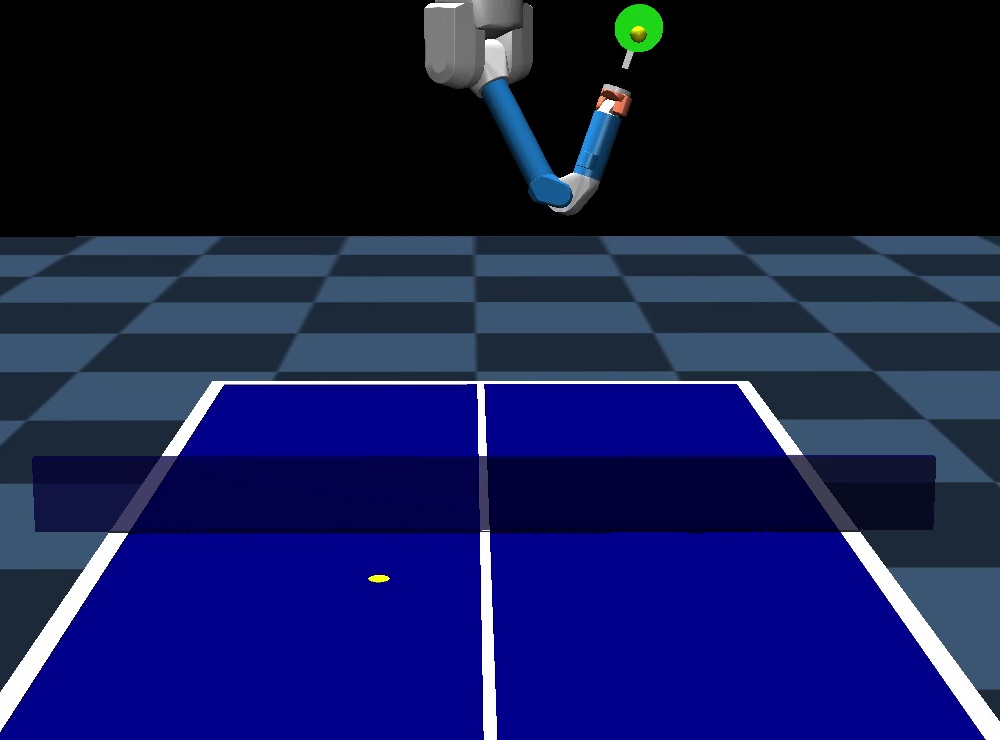}}
       \label{}
   \end{minipage}\hfill

   \begin{minipage}[b]{0.2\textwidth}
        \centering
       \resizebox{\textwidth}{!}{\includegraphics{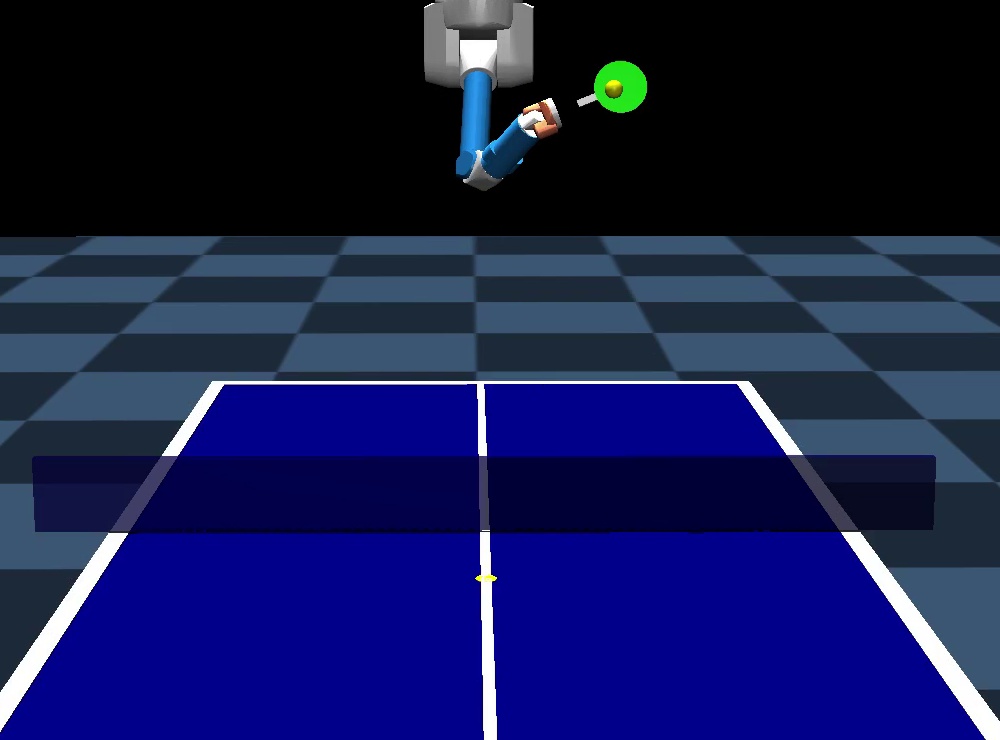}}
       \label{}
    \end{minipage}\hfill
   \begin{minipage}[b]{0.2\textwidth}
        \centering
       \resizebox{\textwidth}{!}{\includegraphics{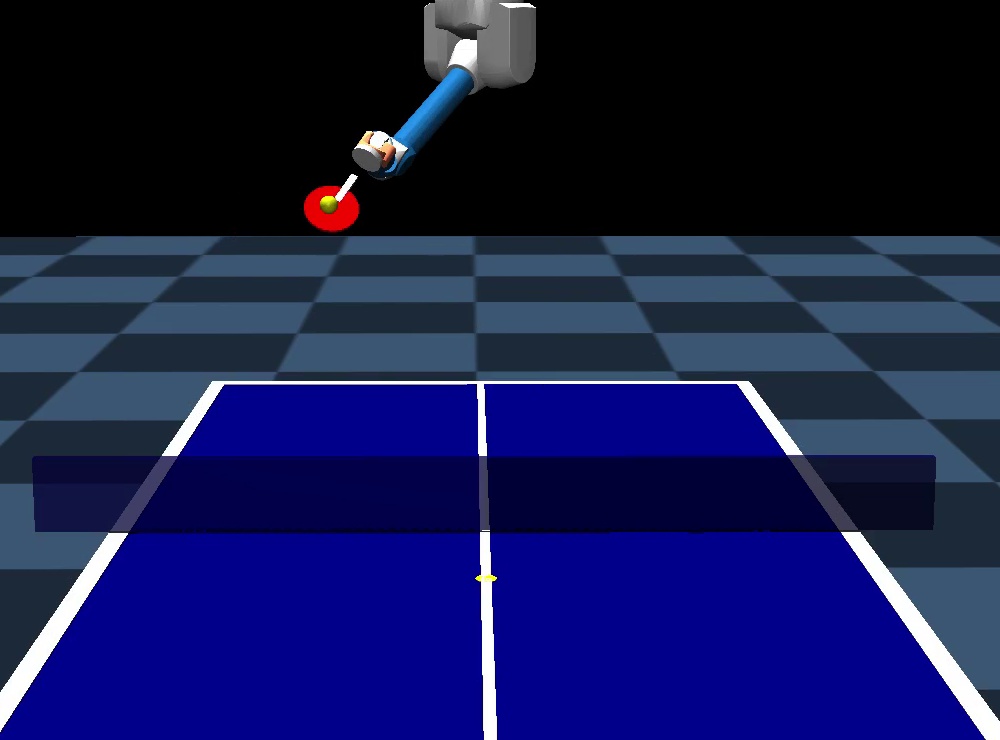}}
       \label{}
    \end{minipage}\hfill
    \begin{minipage}[b]{0.2\textwidth}
        \centering
       \resizebox{\textwidth}{!}{\includegraphics{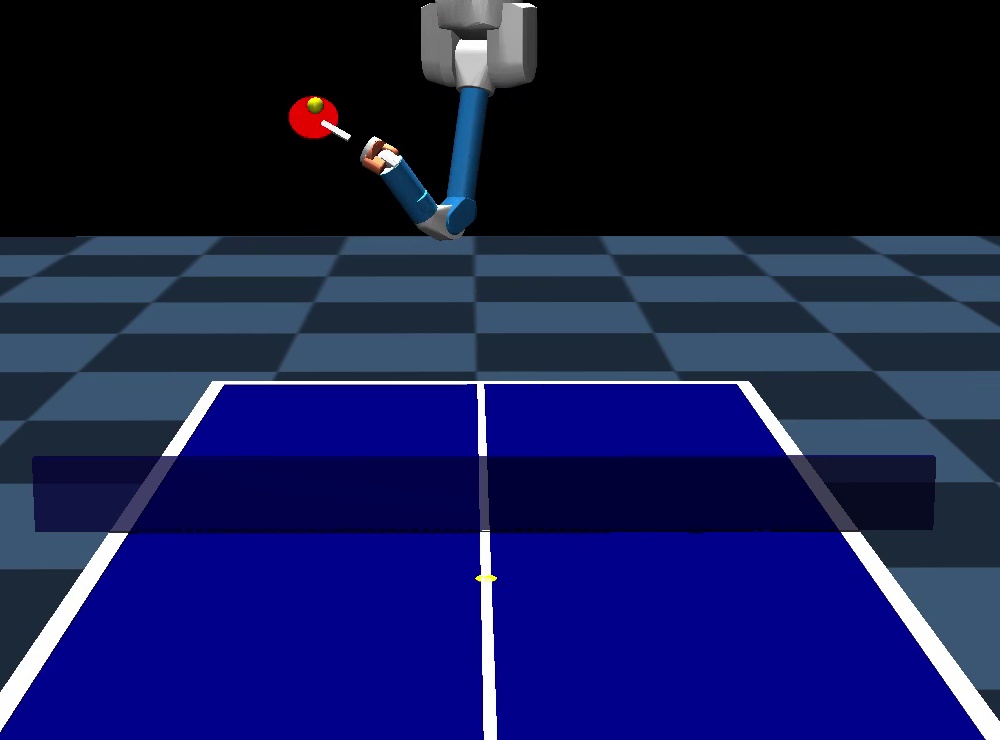}}
       \label{}
   \end{minipage}\hfill
   \begin{minipage}[b]{0.2\textwidth}
        \centering
       \resizebox{\textwidth}{!}{\includegraphics{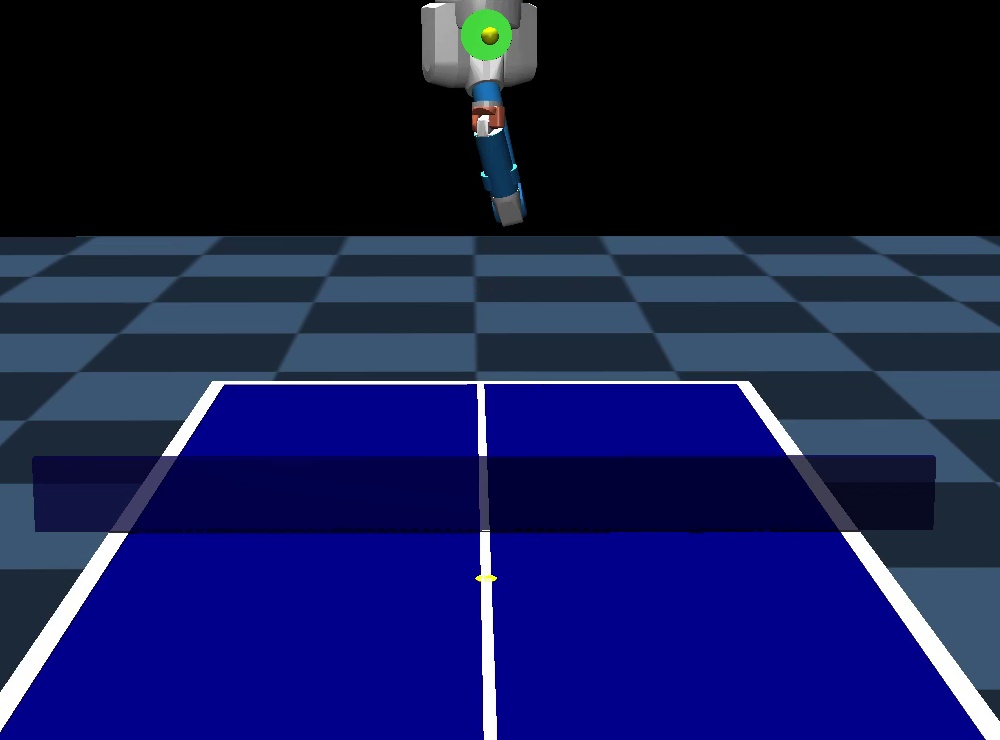}}
       \label{}
   \end{minipage}\hfill
   \begin{minipage}[b]{0.2\textwidth}
        \centering
       \resizebox{\textwidth}{!}{\includegraphics{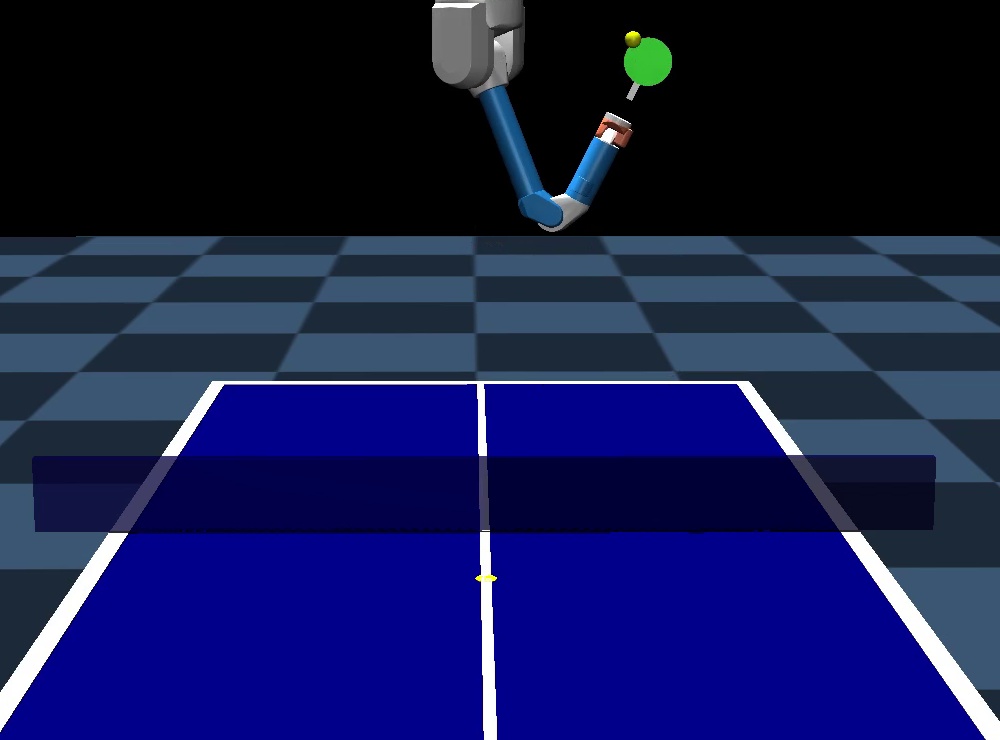}}
       \label{}
   \end{minipage}\hfill
   \caption{Di-SkilL's \textbf{Diverse Skills} for the \textbf{TT-H} task. We fixed the ball's desired landing position and varied the serving landing position and the ball's initial velocity. Di-SkilL can return the ball in different striking types. Note that each row represents a different desired ball landing position.}
   \label{fig::exps::diverse_strikes_appendix}
\end{figure*}

We provide additional diverse skills to the Box Pushing Obstacle task in Fig. \ref{fig::exps::diverse_box_trajectories_appendix}. In Fig. \ref{fig::exps::diverse_strikes_appendix} we provide additional diverse strikes to fixed ball's desired landing positions on the \textbf{TT-H} task.
\begin{figure*}[h]
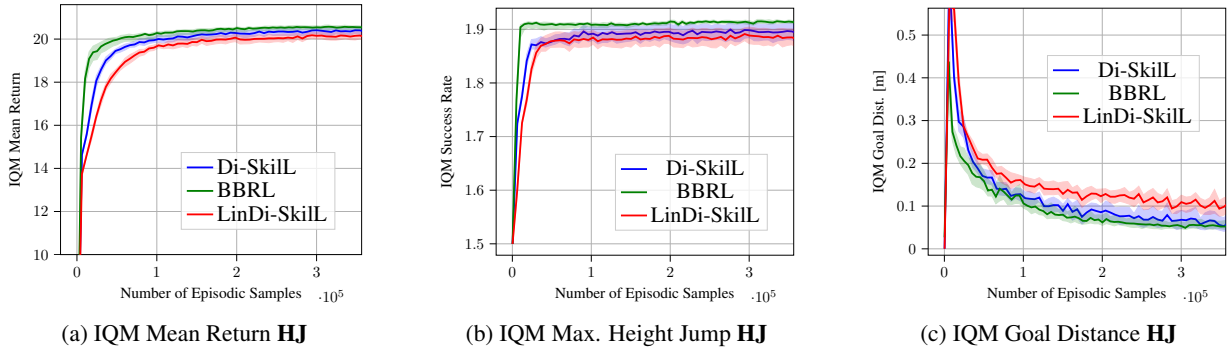

    \centering
   \begin{minipage}[b]{0.33\textwidth}
        \centering
       \resizebox{0.85\textwidth}{!}{\input{figures/experiments/hj/mean_rewards_appendix}}
       \subcaption[]{IQM Mean Return {\bf HJ}}
       \label{fig::exps::hj_rew_appendix}
    \end{minipage}\hfill
    \begin{minipage}[b]{0.33\textwidth}
        \centering
       \resizebox{0.85\textwidth}{!}{\input{figures/experiments/hj/max_height}}
       \subcaption[]{IQM Max. Height Jump {\bf HJ}}
       \label{fig::exps::hj_max_height_appendix}
    \end{minipage}\hfill
    \begin{minipage}[b]{0.33\textwidth}
        \centering
       \resizebox{0.85\textwidth}{!}{\input{figures/experiments/hj/goal_dist}}
       \subcaption[]{IQM Goal Distance {\bf HJ}}
       \label{fig::exps::hj_goal_dist_appendix}
    \end{minipage}\hfill
   \caption{Additional Analysis of the Hopper Jump (HJ) task.
   }
   \label{fig::exps:hj_appendix}
\end{figure*}

Furthermore, we analyze Di-SkilL's performance on the hopper jump task in more detail. In Fig. \ref{fig::exps::hj_rew_appendix} we observe that the mean return is on par with BBRL, similar to the achieved goal distance in Fig. \ref{fig::exps::hj_goal_dist_appendix}. However, there is a small gap in the max height, where BBRL jumps slightly higher (see Fig. \ref{fig::exps::hj_max_height_appendix}. Given that the mean return is on par, one would expect that the maximum jump height is on par as well. However, Di-SkilL optimizes the remaining terms in the objective of the hopper jump task such as the healthy reward (see Appendix \ref{sec::Appendix:ExpDetails}), which explains this gap.

\newpage
\section{Hyperparameters}
We list the hyperparameters for all algorithms on all environments in the following tables.
\begin{table*}[h!]
    \centering
    \begin{tabular}{|c|c|}
        \hline  
        add component every iteration & 1000 \\
        \hline
        fine tune all components every iteration & 50 \\
        \hline
        number component adds & 1\\
        \hline
        number initial components & 1\\
        \hline
        number total components & 20\\
        \hline
        number traj. samples per component per iteration & 200\\
        \hline
        $\alpha$ & 0.0001\\
        \hline
        $\beta$ & 0.5\\
        \hline
        expert KL-bound & 0.01\\
        \hline
        context KL-bound & 0.01\\
        \hline
    \end{tabular}
    \caption{Hyperparameters for SVSL on TT}
    \label{tab:hypparams_svsl}
\end{table*}

\begin{table*}[h!]
    \centering
    \begin{tabular}{|c|c|c|}
    \hline
         & \textbf{Di-SkilL} & \textbf{BBRL} \\
         \hline
        critic activation & tanh & tanh \\
        \hline
        hidden sizes critic & [8,8] & [32, 32]  \\
        \hline
        initialization & orthogonal & orthogonal\\
        \hline
        lr critic& 0.0003 & 0.0003 \\
        \hline
         optimizer critic & adam & adam\\
         \hline
        ciritc epochs  &100  & 100 \\
        \hline
        activation context distribution & tanh & -- \\
        \hline
        epochs context distribution & 100 & -- \\
        \hline
        hidden sizes context distr & [16,16] & -- \\
        \hline
        initialization & orthogonal & -- \\
        \hline
        lr context distribution & 0.0001 &-- \\
        \hline
        optimizer context distr & adam & -- \\
        \hline
        batch size per component& 50 & 209 \\
        \hline
        number samples from environment distribution & 5000 & --\\
        \hline
        number samples per component & 50 & 209\\
        \hline
        normalize advantages & True & True\\
        \hline
        expert activateion & tanh & tanh \\
        \hline
        epochs & 100 & 100 \\
        \hline
        hidden sizes expert & [64] & [32]\\
        \hline
        lr policy & 0.0003 & 0.0003\\
        \hline
        covariance type & full & full \\
        \hline
        alpha & 0.001 & --\\
        \hline
        beta & 4 & --\\
        \hline
        number components & 5 & --\\
        \hline
        covariance bound & 0.005 & 0.001\\
        \hline
        mean bound &0.05 & 0.05\\
        \hline
        projection type & KL & KL \\
        \hline
        trust region coefficient & 100 & 25\\
        \hline
    \end{tabular}
    \caption{Hyperparameters for Di-SkilL and BBRL on TT.}
    \label{tab:hyperparams_ablation_bbrl_ours}
\end{table*}

\begin{table*}[h!]
    \centering
    \begin{tabular}{|c|c|c|c|c|}
    \hline
         & \textbf{Di-SkilL} & \textbf{BBRL} & \textbf{LinDi-SkilL} & \textbf{PPO}\\
        \hline
        critic activation & tanh & tanh & tanh & tanh\\
        \hline
        hidden sizes critic & [32,32] & [32, 32] & [32, 32] & [32, 32]\\
        \hline
        initialization & orthogonal & orthogonal & orthogonal & orthogonal\\
        \hline
        lr critic& 0.0003 & 0.0003 & 0.0003 & 0.0003\\
         \hline
         optimizer critic & adam & adam & adam & adam\\
        \hline
        ciritc epochs  &100  & 100 & 100 & 10\\
        \hline
        activation context distribution & tanh & -- & tanh & --\\
        \hline
        epochs context distribution & 100 & -- & 100 & --\\
        \hline
        hidden sizes context distr & [16,16] & -- & [16, 16] & --\\
        \hline
        initialization & orthogonal & -- & orthogonal & --\\
        \hline
        lr context distribution & 0.0001 &-- & 0.0001 & --\\
        \hline
        optimizer context distr & adam & -- & adam & --\\
        \hline
        batch size per component& 25 & 240 & 25 & 512 (32 minibatches)\\
        \hline
        number samples from environment distribution & 5000 & --& 5000 & --\\
        \hline
        number samples per component & 25 & 240 & 25 & 16384\\
        \hline
        normalize advantages & True & True & True & True\\
        \hline
        expert activateion & tanh & tanh & -- & tanh\\
        \hline
        epochs & 100 & 100 & 100 & 10\\
        \hline
        hidden sizes expert & [32,32] & [64,64] & -- & [32, 32]\\
        \hline
        lr policy & 0.0003 & 0.0003 & 0.0003 & 0.0003\\
        \hline
        covariance type & full & full & full & diagonal\\
        \hline
        alpha & 0.01 & -- & 0.01 & --\\
        \hline
        beta & 8 & -- & 8& --\\
        \hline
        number components & 10 & -- & 10 & --\\
        \hline
        covariance bound & 0.001 & 0.005 & 0.0005 & --\\
        \hline
        mean bound &0.05 & 0.05 & 0.05 & --\\
        \hline
        projection type & KL & KL & KL & --\\
        \hline
        trust region coefficient & 100 & 25 & 100 & --\\
        \hline
        discount factor & 1 & 1 & 1& 1\\
        \hline
    \end{tabular}
    \caption{Hyperparameters for Di-SkilL, BBRL, LinDi-SkilL, and PPO on 5LR. We used all code-level optimization \cite{engstrom2020implementation} needed for PPO. The implementation is based on the source code from \cite{otto2021differentiable}.}
    \label{tab:Reacher_bbrl_ours}
\end{table*}

\begin{table*}[h!]
    \centering
    \begin{tabular}{|c|c|c|c|}
    \hline
         & \textbf{Di-SkilL} & \textbf{BBRL} & \textbf{LinDi-SkilL}\\
         \hline
        critic activation & tanh & tanh & tanh \\
        \hline
        hidden sizes critic & [8,8] & [32, 32] & [8,8] \\
        \hline
        initialization & orthogonal & orthogonal & orthogonal\\
        \hline
        lr critic& 0.0003 & 0.0003 & 0.0003\\
        \hline
         optimizer critic & adam & adam & adam\\
         \hline
        ciritc epochs  &100  & 100 & 100\\
        \hline
        activation context distribution & tanh & -- & tanh\\
        \hline
        epochs context distribution & 100 & -- & 100\\
        \hline
        hidden sizes context distr & [16,16] & -- & [16, 16\\
        \hline
        initialization & orthogonal & -- & orthogonal\\
        \hline
        lr context distribution & 0.0001 &-- &0.0001\\
        \hline
        optimizer context distr & adam & -- & adam\\
        \hline
        batch size per component& 50 & 209 &50\\
        \hline
        number samples from environment distribution & 5000 & -- & 5000\\
        \hline
        number samples per component & 50 & 209 & 50\\
        \hline
        normalize advantages & True & True & True\\
        \hline
        expert activateion & tanh & tanh & --\\
        \hline
        epochs & 100 & 100\\
        \hline
        hidden sizes expert & [128] & [32,32] & --\\
        \hline
        lr policy & 0.0003 & 0.0003 & 0.0003\\
        \hline
        covariance type & full & full & full \\
        \hline
        alpha & 0.001 & -- & 0.001\\
        \hline
        beta & 0.5 & -- & 0.5\\
        \hline
        number components & 10 & -- & 10\\
        \hline
        covariance bound & 0.005 & 0.0005& 0.001\\
        \hline
        mean bound &0.05 & 0.05 & 0.05\\
        \hline
        projection type & KL & KL & KL\\
        \hline
        trust region coefficient & 100 & 25 &100\\
        \hline
    \end{tabular}
    \caption{Hyperparameters for Di-SkilL, BBRL, and LinDi-SkilL for the Hard Table Tennis Task (TT-H).}
    \label{tab:hyperparams_extended_TT_bbrl_ours}
\end{table*}

\begin{table*}[h!]
    \centering
    \begin{tabular}{|c|c|c|c|}
    \hline
         & \textbf{Di-SkilL} & \textbf{BBRL} & \textbf{LinDi-SkilL}\\
         \hline
        critic activation & tanh & tanh & tanh \\
        \hline
        hidden sizes critic & [64,64] & [64, 64] & [64,64] \\
        \hline
        initialization & orthogonal & orthogonal & orthogonal\\
        \hline
        lr critic& 0.0001 & 0.0001 & 0.0001\\
        \hline
         optimizer critic & adam & adam & adam\\
         \hline
        ciritc epochs  &100  & 100 & 100\\
        \hline
        activation context distribution & tanh & -- & tanh\\
        \hline
        epochs context distribution & 100 & -- & 100\\
        \hline
        hidden sizes context distr & [16,16] & -- & [16, 16]\\
        \hline
        initialization & orthogonal & -- & orthogonal\\
        \hline
        lr context distribution & 0.0001 &-- &0.0001\\
        \hline
        optimizer context distr & adam & -- & adam\\
        \hline
        batch size per component& 80 & 200 &80\\
        \hline
        number samples from environment distribution & 1000 & -- & 1000\\
        \hline
        number samples per component & 80 & 200 & 80\\
        \hline
        normalize advantages & True & True & True\\
        \hline
        expert activateion & tanh & tanh & --\\
        \hline
        epochs & 100 & 100 &100\\
        \hline
        hidden sizes expert & [32, 32] & [32,32] & --\\
        \hline
        lr policy & 0.0003 & 0.0003 & 0.0003\\
        \hline
        covariance type & full & full & full \\
        \hline
        alpha & 0.01 & -- & 0.01\\
        \hline
        beta & 8 & -- & 8\\
        \hline
        number components & 3 & -- & 3\\
        \hline
        covariance bound & 0.005 & 0.05& 0.005\\
        \hline
        mean bound &0.05 & 0.1 & 0.05\\
        \hline
        projection type & KL & KL & KL\\
        \hline
        trust region coefficient & 100 & 25 &100\\
        \hline
    \end{tabular}
    \caption{Hyperparameters for Di-SkilL, BBRL, and LinDi-SkilL for the Hopper Jump Task (HJ).}
    \label{tab:hyperparams_hopper_jump}
\end{table*}

\begin{table*}[h!]
    \centering
    \begin{tabular}{|c|c|c|c|c|}
    \hline
         & \textbf{Di-SkilL} & \textbf{BBRL} & \textbf{LinDi-SkilL} & \textbf{PPO}\\
        \hline
        critic activation & tanh & tanh & tanh & tanh\\
        \hline
        hidden sizes critic & [32,32] & [32, 32] & [32, 32] & [256, 256] \\
        \hline
        initialization & orthogonal & orthogonal & orthogonal & orthogonal\\
        \hline
        lr critic& 0.0003 & 0.0003 & 0.0003 & 0.0001\\
         \hline
         optimizer critic & adam & adam & adam & adam\\
        \hline
        ciritc epochs  &100  & 100 & 100 & 10\\
        \hline
        activation context distribution & tanh & -- & tanh & --\\
        \hline
        epochs context distribution & 100 & -- & 100 & --\\
        \hline
        hidden sizes context distr & [16,16] & -- & [16, 16] & --\\
        \hline
        initialization & orthogonal & -- & orthogonal & --\\
        \hline
        lr context distribution & 0.0001 &-- & 0.0001 & --\\
        \hline
        optimizer context distr & adam & --& adam  & --\\
        \hline
        batch size per component& 50 &500 & 50 & 410 (40 minibatches)\\
        \hline
        number samples from environment distribution & 5000 & -- & 5000 & --\\
        \hline
        number samples per component & 50 & 500 & 50 & 16384\\
        \hline
        normalize advantages & True & True & True& True \\
        \hline
        expert activateion & tanh & tanh & --& tanh\\
        \hline
        epochs & 100 & 100 & 100 & 10\\
        \hline
        hidden sizes expert & [64,64] & [64,64 & -- & [256, 256] \\
        \hline
        lr policy & 0.0003 & 0.0003 & 0.0003 & 0.0001\\
        \hline
        covariance type & full & full & full & diagonal\\
        \hline
        alpha & 0.01 & -- & 0.0001 & --\\
        \hline
        beta & 64 & -- & 64 &--\\
        \hline
        number components & 10 & -- & 10 & --\\
        \hline
        covariance bound & 0.005 & 0.0005 & 0.001 & --\\
        \hline
        mean bound &0.05 & 0.05 & 0.05 & --\\
        \hline
        projection type & KL & KL & KL & --\\
        \hline
        trust region coefficient & 100 & 25 & 100 & --\\
        \hline
        discount factor & 1 & 1 & 1& 1\\
        \hline
    \end{tabular}
    \caption{Hyperparameters for Di-SkilL, BBRL, LinDi-SkilL, and PPO for Box Pushing Obstacle task (BPO). We used all code-level optimization \cite{engstrom2020implementation} needed for PPO. The implementation is based on the source code from \cite{otto2021differentiable}.}
    \label{tab:BoxP_bbrl_ours}
\end{table*}

\begin{table*}[h!]
    \centering
    \begin{tabular}{|c|c|c|c|}
    \hline
         & \textbf{Di-SkilL} & \textbf{BBRL} & \textbf{LinDi-SkilL}\\
        \hline
        critic activation & tanh & tanh & tanh \\
        \hline
        hidden sizes critic & [32,32] & [32, 32] & [32, 32] \\
        \hline
        initialization & orthogonal & orthogonal & orthogonal\\
        \hline
        lr critic& 0.0003 & 0.0003 & 0.0003\\
         \hline
         optimizer critic & adam & adam & adam\\
        \hline
        ciritc epochs  &100  & 100 & 100\\
        \hline
        activation context distribution & tanh & -- & tanh \\
        \hline
        epochs context distribution & 100 & -- & 100\\
        \hline
        hidden sizes context distr & [16,16] & -- & [16, 16]\\
        \hline
        initialization & orthogonal & -- & orthogonal\\
        \hline
        lr context distribution & 0.0001 &-- & 0.0001\\
        \hline
        optimizer context distr & adam & -- & adam\\
        \hline
        batch size per component& 50 & 500 & 50\\
        \hline
        number samples from environment distribution & 5000 & -- & 5000\\
        \hline
        number samples per component & 50 & 500 & 50\\
        \hline
        normalize advantages & True & True & True\\
        \hline
        expert activateion & tanh & tanh & --\\
        \hline
        epochs & 100 & 100 & 100\\
        \hline
        hidden sizes expert & [64,64] & [128,128] & --\\
        \hline
        lr policy & 0.0003 & 0.0003 & 0.0003\\
        \hline
        covariance type & full & full & full\\
        \hline
        alpha & 0.0001 & -- & 0.0001\\
        \hline
        beta & 1 & -- & 1\\
        \hline
        number components & 10 & -- & 10\\
        \hline
        covariance bound & 0.005 & 0.001& 0.001\\
        \hline
        mean bound &0.05 & 0.05 &0.01\\
        \hline
        projection type & KL & KL & KL\\
        \hline
        trust region coefficient & 100 & 25 & 100\\
        \hline
    \end{tabular}
    \caption{Hyperparameters for Di-SkilL, BBRL, and LinDi-SkilL for the mini golf task.}
    \label{tab:minigolf_bbrl_ours}
\end{table*}

\end{document}